\pdfoutput=1

\documentclass[11pt]{article}

\usepackage[final]{acl}

\usepackage{times}
\usepackage{latexsym}

\usepackage[T1]{fontenc}

\usepackage[utf8]{inputenc}

\usepackage{microtype}

\usepackage{inconsolata}

\usepackage{graphicx}
\usepackage{xspace}
\usepackage{booktabs}
\usepackage{adjustbox}
\usepackage{amssymb}
\usepackage{colortbl}
\usepackage{xcolor}
\usepackage{longtable}

\usepackage[most]{tcolorbox}
\usepackage{colortbl}
\newtcolorbox{prompt}[2][]{
    colback=white,
    colframe=gray!45,
    fonttitle=\bfseries,
    coltitle=black,
    title=#2,
    #1,breakable
}
\usepackage{cleveref}
\usepackage{arydshln}

\newcommand{\datasetName}{\textsc{Stark}\xspace}
\newcommand{\datasetEmoji}{\includegraphics[height=.9em,trim=0 .4em 0 0]{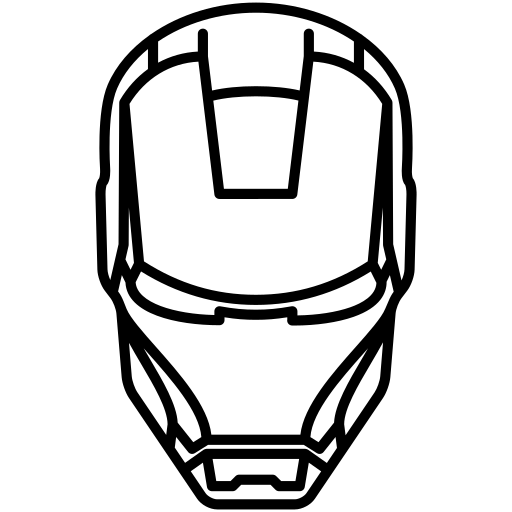}}
\newcommand{\dataset}{\datasetEmoji \xspace \datasetName \xspace}

\newcommand{\frameworkName}{\textsc{Mcu}\xspace}

\newcommand{\modelName}{\textsc{Ultron}\xspace}
\newcommand{\modelEmoji}{\includegraphics[height=.9em,trim=0 .4em 0 0]{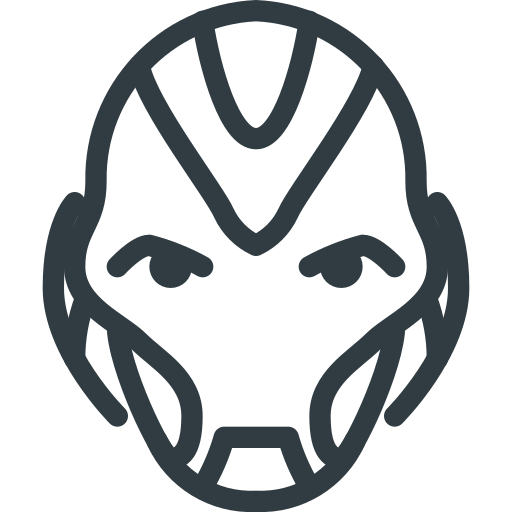}}
\newcommand{\model}{\modelEmoji \xspace \modelName}

\newcommand{\planExecute}{Plan-and-Execute\xspace}

\newcommand{\driber}{\textsc{DribeR}\xspace}

\newcommand{\Routine}{Routine\xspace}
\newcommand{\Goal}{Goal\xspace}
\newcommand{\Experience}{Experience\xspace}
\newcommand{\Relationship}{Relationship\xspace}
\newcommand{\Characteristic}{Characteristic\xspace}

\newcommand{\system}{System Message}
\newcommand{\user}{Instruction}
\newcommand{\userRoutine}{Instruction for \Routine Relation}
\newcommand{\userGoal}{Instruction for \Goal Relation}
\newcommand{\userExperience}{Instruction for \Experience Relation}
\newcommand{\userRelationship}{Instruction for \Relationship Relation}
\newcommand{\userCharacteristic}{Instruction for \Characteristic Relation}

\newcommand{\grayrow}{\rowcolor[gray]{0.9}}

\newcommand{\eg}{e.g.,\xspace}
\newcommand{\ie}{i.e.,\xspace}

\crefname{section}{§}{§§}

\title{\dataset: Social Long-Term Multi-Modal Conversation with Persona Commonsense Knowledge}

\author{
Young-Jun Lee \textsuperscript{\rm 1} \quad
Dokyong Lee \textsuperscript{\rm 2} \quad
Junyoung Youn \textsuperscript{\rm 2} \quad
Kyeongjin Oh \textsuperscript{\rm 2} \quad \\
\textbf{Byungsoo Ko} \textsuperscript{\rm 1} \quad
\textbf{Jonghwan Hyeon} \textsuperscript{\rm 1} \quad
\textbf{Ho-Jin Choi} \textsuperscript{\rm 1} 
\\
\textsuperscript{\rm 1} School of Computing, KAIST \quad
\textsuperscript{\rm 2} KT Corporation
\\
\texttt{\{yj2961, jonghwanhyeon, hojinc\}@kaist.ac.kr} \quad
\texttt{Kobiso62@gmail.com} \\
\texttt{\{dokyong.lee, junyoung.youn, kyeong-jin.oh\}@kt.com}
}

\begin{document}
\maketitle

\begin{abstract}

Humans share a wide variety of images related to their personal experiences within conversations via instant messaging tools. However, existing works focus on (1) image-sharing behavior in singular sessions, leading to limited long-term social interaction, and (2) a lack of personalized image-sharing behavior. In this work, we introduce \dataset, a large-scale long-term multi-modal conversation dataset that covers a wide range of social personas in a multi-modality format, time intervals, and images. To construct \datasetName automatically, we propose a novel multi-modal contextualization framework, \frameworkName, that generates long-term multi-modal dialogue distilled from ChatGPT and our proposed \planExecute image aligner. Using our \datasetName, we train a multi-modal conversation model, \model 7B, which demonstrates impressive visual imagination ability. Furthermore, we demonstrate the effectiveness of our dataset in human evaluation. We make our source code and dataset publicly available~\footnote{\url{https://stark-dataset.github.io/}}.
    
\end{abstract}

\section{Introduction} \label{sec:intro}

For decades, the development of empowering human-computer interaction has been steadily advancing across various domains (\eg social dialogue~\cite{zhou2023sotopia}, writing~\cite{lee2022coauthor,han2023recipe}), multifaceted ingredients (\eg affective user's state~\cite{hudlicka2003feel}, multi-perspective~\cite{kammersgaard1988four}, multiple social skills~\cite{yang2024social}) and multi-modality~\cite{jaimes2007multimodal} with the goal of increasing human satisfaction and engagement. 
To strengthen the interaction in a practicable real scenario, recent system~\cite{shin2023introbot} have adopted the \textit{image-sharing behavior}~\cite{lobinger2016photographs}, an interaction frequently occurring via instant messaging tools, interpreting it as a communicative practice. Consequently, previous studies have proposed multi-modal dialogue datasets through various methods, including crowd-sourcing~\cite{zang2021photochat}, social media~\cite{feng2022mmdialog}, and distillation from large language models (LLMs)~\cite{lee2024dialogcc,aboutalebi2024magid,maharana2024evaluating}.

\begin{figure*}[t]
    \centering
    \includegraphics[width=\linewidth]{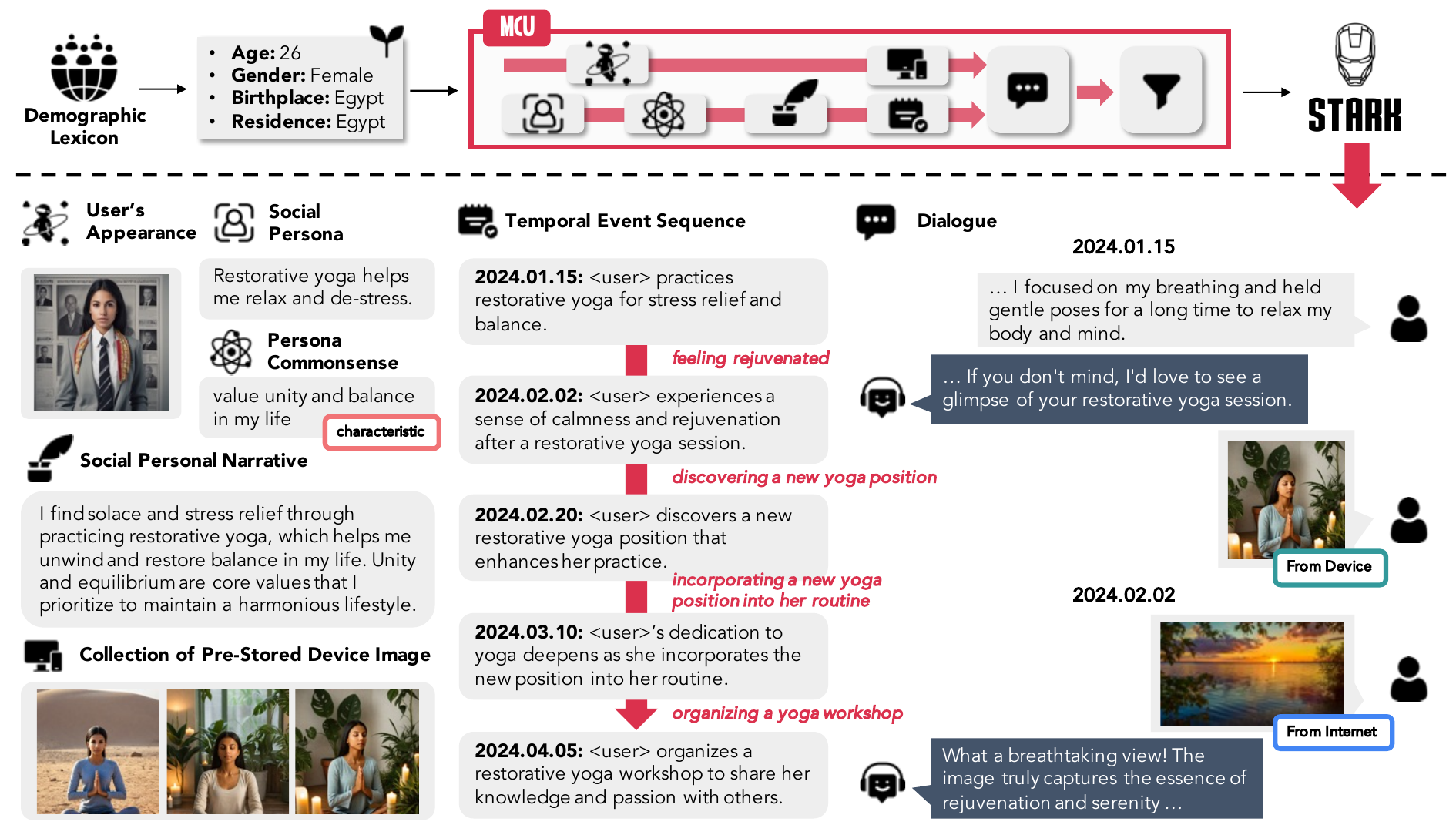}
    \caption{An overview of \frameworkName and an example of \dataset. At the top, our framework takes basic demographic information (\ie age, gender, birthplace, residence) and generates a long-term multi-modal conversation. At the bottom, our \datasetName includes various information such as user's appearance, social persona, persona commonsense, personal narrative, a collection of pre-stored device images, temporal event sequences, and multi-modal dialogue. In this figure, a short sentence between two events indicates the user's episodic experience between those events (\eg ``\textit{felling rejuvenated}'').}
    \label{main_fig:pipeline}
    \vspace{-1em}
\end{figure*}

However, existing datasets are limited in their representation of \textit{personalized image-sharing behavior} over extended periods beyond a singular time (\eg a few hours, days, weeks), preventing trained multi-modal dialogue models from seamlessly communicating with users in real-world human-bot interactive scenarios. For example, as shown in Figure~\ref{main_fig:pipeline}, depending on who is the user (\ie human's appearance), there is a user's appearance and user's personal experience inside the shared image. Nevertheless, existing datasets regarding multi-modal dialogue do not consider multi-modality persona information (in Table~\ref{main_tab:dataset_stat}).

To address this issue, we first introduce a large-scale \textbf{S}ocial long-\textbf{T}erm multi-mod\textbf{A}l conve\textbf{R}sation dataset with persona commonsense \textbf{K}nowledge, \dataset, covering a wide variety of social personal dynamics (\ie \textit{demographics}, \textit{personal experience}), more realistic time intervals, and personalized images. 
To construct \datasetName, we propose a novel framework, \frameworkName, that distills long-term multi-modal dialogue from a large language model (LLM)~\footnote{In this work, we use ChatGPT~\cite{chatgpt}, but our proposed framework can work with any large language models, such as LLaMA-3~\cite{llama3modelcard}.} and our proposed \planExecute image aligner, powered by a personalized text-to-image generative model, image database retrieval, and web search, as shown in Figure~\ref{main_fig:pipeline}. As a result of being grounded on various personal dynamics over a long period, \datasetName contains more personalized multi-modal conversation dataset. In addition, even though \datasetName is automatically constructed, \datasetName show higher preferred quality compared to other multi-modal conversation datasets (\cref{sec:human_eval}). With our \dataset dataset, we build a multi-modal conversation model, \model 7B, which is fine-tuned model on top of recent multi-modal language model~\cite{lee2024meteor}. As a result, \modelName achieves significant performance on dialogue-to-image retrieval task which implies the effectiveness of our dataset.

In summary, our main contributions are as follows: 1) We propose the first large-scale social long-term multi-modal conversation dataset, \dataset, covering the personalized image-sharing behavior. 2) To construct \datasetName, we propose a multi-modal contextualization framework, \frameworkName, that generate a multi-modal dialogue over a time period by only providing basic demographic information. 3) Using our dataset, we build a multi-modal converstation model, \model 7B. 4) Through extensive experiments, we demonstrate the effectiveness and reliability of our dataset and framework in human evaluation and dialogue-to-image retrieval tasks.

\section{Related Work} \label{sec:related_work}

\paragraph{Multi-Modal Dialogue Dataset.}

In the dynamic field of multi-modal dialogue, most previous studies are categorized into two primary groups: those where the image is grounded at the beginning of the dialogue and those where the image is shared during the dialogue. The image-grounded dialogue task aims to answer questions~\cite{antol2015vqa,das2017visual,seo2017visual,kottur2019clevr} or generate natural conversations~\cite{mostafazadeh2017image,shuster2018image,meng2020openvidial,wang2021openvidial,zheng2021mmchat} about given images by considering the comprehensive multi-modal persona information~\cite{ahn2023mpchat}. However, in our daily conversations, we often share images relevant to the context of the dialogue via instant messaging tools. Inspired by this behavior, recently proposed image-sharing dialogue datasets have been constructed through crowd-sourcing~\cite{zang2021photochat}, social media~\cite{feng2022mmdialog}, image-text matching model~\cite{lee2021constructing}, or annotating image-sharing moments~\cite{lee2024dialogcc,aboutalebi2024magid} through large language models (LLMs). These datasets boast impressive quality and image diversity. However, they are confined to a single session, which hinders the ability of trained models to maintain continuous conversations with users and potentially disrupts the interaction between the user and the AI assistant.

\paragraph{Constructing Dialogue Dataset using Large Language Models.}
To effectively address the pervasive issue of data scarcity, several innovative studies have leveraged large language models (LLMs) to construct diverse and scalable dialogue datasets. These efforts encompass personalized dialogue~\cite{lee2022personachatgen,jandaghi2023faithful}, multi-turn dialogue for prosocial behavior~\cite{kim2022prosocialdialog}, million-scale social dialogue~\cite{kim2022soda} by contextualizing rich social commonsense knowledge from a comprehensive knowledge graph~\cite{west2021symbolic}, theory-of-mind (ToM) related multi-party dialogue~\cite{kim2023fantom}, multi-hop reasoning over dialogue~\cite{chae2023dialogue}, long-term dialogue~\cite{jang2023conversation,zhang2023mind}, and multi-modal dialogue~\cite{lee2024dialogcc,aboutalebi2024magid}. Recently, a novel multi-modal dialogue dataset~\cite{maharana2024evaluating} encompassing multiple sessions has been proposed. However, this particular dataset is designed primarily as an evaluation benchmark, thus complicating the development of an adequate multi-modal dialogue model. Furthermore, this dataset does not prioritize multi-modality in the context of personalization during image-sharing interactions. In this work, we are excited to introduce the concept of personalized multi-modal conversations over extended time intervals, meticulously considering the dynamic nature of personal interactions.

\section{\frameworkName: A Multi-Modal Contextualization Framework for Conversation Distillation}

Inspired from recent study~\cite{kim2022soda}, we propose a \frameworkName, a multi-modal contextualization framework for distilling long-term multi-modal dialogue from combination of large language model (LLM) and our proposed \planExecute image aligner.
Specifically, \frameworkName consists of eight steps: (1) Generating social persona attribute based on the collection of demographics (\ie age, gender, birthplace, residence) (\cref{sec:persona}), (2) generating virtual human face based on the demographic (\cref{sec:human_face}) (3) generating social persona commonsense knowledge based on the generated social persona sentence and the demographic (\cref{sec:commonsense}), (4) generating a social personal narrative from the commonsense knowledge (\cref{sec:narrative}), grounding on the personal narrative we (5) generate an event sequence (\cref{sec:event}) and (6) generate a collection of pre-stored device images (\cref{sec:device}), (7) generating a multi-modal conversation with multiple sessions over a diverse time period (\cref{sec:conversation}), and (8) aligning a realistic and personalized image to the generated image-sharing moment by leveraging proposed \planExecute image aligner (\cref{sec:planExecute}). The overview of our framework is illustrated in Figure~\ref{main_fig:pipeline}. In all steps of our framework, we use ChatGPT~\cite{chatgpt} (\ie \texttt{gpt-3.5-turbo-0125}) as our LLM. All prompt templates used in our framework are presented in Appendix~\ref{supp_sec:prompt_template}.

\subsection{Motivation Behind Grounded on Demographic}\label{sec:demographic}

Social interactions are a core component of human life, facilitated primarily through conversation~\cite{myllyniemi1986conversation}. These interactions often involve sharing personal experiences, which can be abstracted into narratives or scripts~\cite{mar2008function}. We posit that these personal experiences are highly dependent on the individual's demographic information (\eg age, country), thereby affect the general topic of interaction socially and culturally. Thus, we start with basic demographic information, age, gender, birthplace, residence.

\subsection{Social Persona}\label{sec:persona}

We first randomly sample demographic information (\ie age, gender, birthplace, residence) from a pre-defined demographic lexicon, as detailed in Appendix~\ref{supp_sec:demographic_lexicon}, by referring to previous work~\cite{santy2023nlpositionality}. From the chosen demographic information, we construct a social persona~\footnote{In this work, we regard a persona as a user profile, following the definition of previous work~\cite{lee2022personachatgen}.} in the form of a short sentence for a persona category among 50 predefined persona categories. 
The complete list of persona categories is provided in the Appendix~\ref{supp_sec:persona_category}. 
Additionally, we generate a social persona attribute simultaneously with the social persona sentence. The social persona attribute can be formally represented as a triple ($e_1$, $r$, $e_2$), where $e_1$, $r$, and $e_2$ denote the persona subject, persona category, and persona entity, respectively. The persona entity follows a key-value format. For example, in the social persona attribute of ``I am from London,'' $e_1$ is ``I,'' $r$ is ``location,'' and $e_2$ is ``(city-state, London).'' To save time and reduce costs, we generate 30 persona attributes and sentences given a single demographic information set and a target persona category.

\subsection{Virtual Human Face} \label{sec:human_face}

Since \datasetName should cover personalized image-sharing behavior, we generate a virtual human face using the SDXL-Lightning~\cite{lin2024sdxllightning} model.~\footnote{Unfortunately, we intended to use a more specialized model~\cite{li2024cosmicman} for human face generation; however, this model was not publicly available at the time of data construction, so we opted for the alternative model.} The virtual human face is created based on a predefined human attribute collection from a recent work~\cite{li2024cosmicman}, with the full human attribute information presented in the Appendix~\ref{supp_sec:human_attr_pool}. 
Creating a virtual human face initially allows us to generate personalized images later (in~\cref{sec:planExecute}) with higher quality and more personalized experiences, resulting in significant scores in human evaluation (\cref{sec:human_eval}).

\subsection{Social Persona Commonsense Knowledge}\label{sec:commonsense}

Recent research has introduced a large-scale persona-grounded commonsense knowledge graph called \textsc{PeaCoK}~\cite{gao2023peacok}. This graph is symbolically represented in the form of triples (head, relation, tail), where relation denotes a defined \textit{persona frame} concept, which formalizes five commonsense aspects of persona knowledge: \texttt{characteristics}, \texttt{routines/habits}, \texttt{goals/plans}, \texttt{experiences}, and \texttt{relationships}. This comprehensive knowledge graph encompasses a broad spectrum of persona knowledge at scale.

However, this commonsense knowledge graph has two major limitations: (1) The coverage of persona head value is limited to the \textit{CapableOf} relation (in \textsc{Atomic}$_{20}^{20}$~\cite{hwang2021comet}), which typically encompasses occupation-related sentences (\eg ``I am a programmer,'' ``I am a basketball player''). In reality, persona identity can be expressed through a broad range of information, such as ``I have two dogs'' in terms of possession. (2) The inferred attribute knowledge based on the given commonsense relation varies depending on demographic information. For example, even when providing the same persona head value and the same commonsense relation, the persona commonsense inference will represent distinct meanings based on demographic differences.

To address these limitations, we prompt ChatGPT to infer the persona attribute knowledge considering the user's demographic information and social persona sentence (\cref{sec:persona}), which covers diverse persona categories, for five persona relations. 

\subsection{Personal Narrative}\label{sec:narrative}

\paragraph{Symbolic Form to Sentence Form.} We convert the generated persona commonsense knowledge graphs into simple sentences by applying predefined templates (presented in the Table~\ref{supp_tab:sentence_form}) for each relation. To make the sentences more plausible and natural in terms of world knowledge, we use actual names based on the given birthplace country, selecting from the Top-1K names for each country~\footnote{\url{https://github.com/philipperemy/name-dataset}}.

\vspace{-0.5em}

\paragraph{Sentence Form to Personal Narrative.} Next, we prompt ChatGPT to transform the sentence form into a short personal narrative consisting of two or three sentences with detailed information, following recent work~\cite{kim2022soda}.

\subsection{Temporal Event Sequence}\label{sec:event}

Starting from the generated personal narrative, we prompt ChatGPT to generate a temporal event sequence consisting of multiple sequential events. We prompt ChatGPT to generate time intervals and episodic experiences with operations between two events. There are two types of experience operations: \texttt{add} and \texttt{update}.~\footnote{In our dataset, the ratio of \texttt{add} and \texttt{update} is 82.9\% and 17.1\%, respectively.} If a new experience occurs, it is marked as an \texttt{add} operation. If a previous experience is modified, it is marked as an \texttt{update} operation. 

\subsection{Collection of Pre-Stored Device Images} \label{sec:device}

Before generating multi-modal conversations, we ask ChatGPT to infer the possible image descriptions that might be pre-stored on a user's device (\eg mobile or laptop) based on the personal narrative (\cref{sec:narrative}). This step makes multi-modal conversations more practical and similar to real-world scenarios, such as when a user shares an everyday photo on online social media~\cite{maclean2022instagram}. Specifically, we generate five image descriptions along with corresponding image categories (see the ratio of image categories in \cref{sec:rich_info}). We then generate photo-realistic images using our proposed \planExecute image aligner (see details in \cref{sec:planExecute}).

\begin{figure}[!t]
    \centering
    \includegraphics[width=\columnwidth]{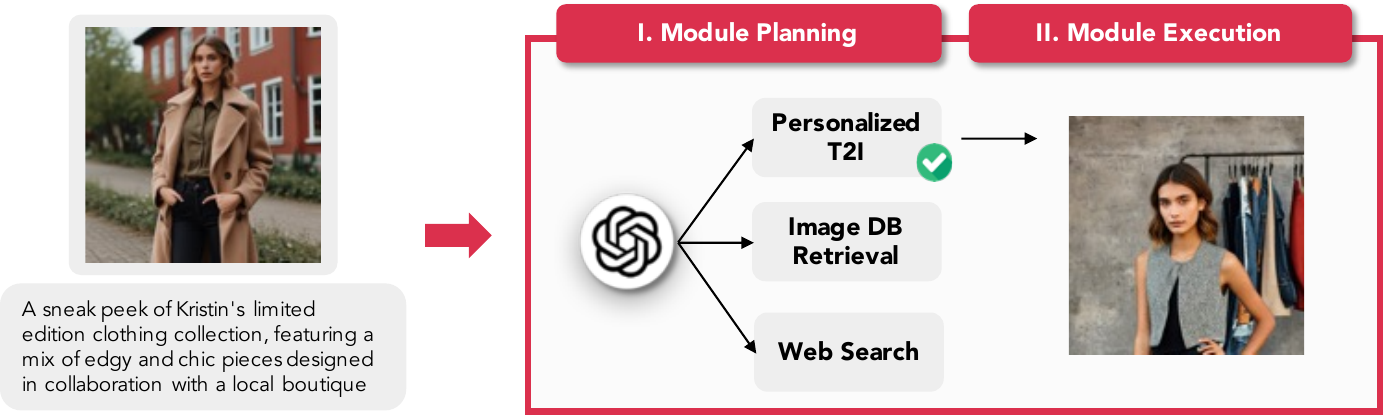}
    \caption{An illustration of our \planExecute image aligner process. }
    \label{main_fig:plan_and_execute}
    \vspace{-1em}
\end{figure}

\subsection{Multi-Modal Conversation}\label{sec:conversation}

In this step, we generate a long-term multi-modal conversation between the user and an AI assistant, utilizing the constructed event sequence (\cref{sec:event}) and the collection of pre-stored device image descriptions (\cref{sec:device}). Since each episode consists of multiple session dialogues, we generate each session sequentially. Concretely, the second session is influenced by useful information (\ie history of events, device images) from the previous session.

\paragraph{Generating Image-Sharing Moment.} Drawing inspiration from recent works~\cite{lee2024dialogcc,lee2023large,aboutalebi2024magid}, we employ ChatGPT to create a multi-modal conversation that includes an image-sharing moment in text format, specifically encompassing \texttt{image description}, \texttt{rationale}, \texttt{image source}, \texttt{keywords}, and \texttt{index of pre-stored image in device}. For \texttt{image source}, to ensure the multi-modal conversations are as realistic and natural as possible, closely mirroring real-life scenarios, we prompt ChatGPT to specify the source of the shared image: either from the internet or from a user's device~\footnote{In our dataset, the ratio of images sourced from the internet to those sourced from a device is 60.8\% to 39.2\%, respectively.}, when describing an image-sharing moment. Furthermore, for \texttt{index of pre-stored image in device}, if the shared image is already part of a collection of pre-stored device image descriptions, we prompt ChatGPT to determine which image description to select.

\paragraph{\planExecute Image Aligner.}\label{sec:planExecute}

Since \datasetName is designed to include personalized image-sharing behavior over an extended period, users can share photos that reflect their personal experiences. For example, a user might share a photo with the description, ``\textit{I visited the Eiffel Tower last week}'', which includes an image of them in front of the Eiffel Tower. Additionally, users can share non-human-centric photos, such as \textit{``a meal I had yesterday''}, which also conveys personal experiences. Therefore, we need to determine the most appropriate module to synthesize images relevant to the given image descriptions. 

Following recent works related to tool-based AI agents~\cite{shen2024hugginggpt}, as illustrated in Figure~\ref{main_fig:plan_and_execute}, we first conduct module planning to select the most appropriate module based on the given image description by leveraging ChatGPT. The options include a personalized text-to-image generator, image database retrieval, and web search. After selecting the appropriate module, we proceed to execute it. Specifically, if the personalized text-to-image generator is chosen, we utilize the PhotoMaker~\cite{li2023photomaker} model, demonstrating impressive performance in customizing human faces. If image database retrieval is selected, we use the CLIP~\cite{radford2021learning} (\ie \texttt{ViT-L/14@336px}) to retrieve relevant images from prepared source image datasets: CC12M~\cite{changpinyo2021conceptual}, RedCaps12M~\cite{desai2021redcaps}, ChartQA~\cite{masry2022chartqa}, AI2D~\cite{kembhavi2016diagram}, and MathVision~\cite{wang2024measuring}. For an efficient search, we utilize an AutoFaiss~\footnote{\url{https://github.com/criteo/autofaiss}}. We employ Bing Search~\footnote{\url{https://pypi.org/project/icrawler/}} for web search, similar to previous work~\cite{maharana2024evaluating}.

\subsection{Post-processing and Filtering} \label{sec:filtering}

We remove episode conversations that have less than four sessions or more than six sessions (7.1\%); remove duplicate persona attributes (19.8\%). In addition, we remove potentially dangerous and harmful dialogues that need the intervention using Canary~\cite{kim2022prosocialdialog} and unsuitable images using NSFW detector~\footnote{\url{https://huggingface.co/Falconsai/nsfw_image_detection}}. Furthermore, we filter out unaligned images to the generated image descriptions using Pick-a-pic~\cite{kirstain2023pick} score. Finally, we obtain roughly 0.5 M session dialogues in total.

{\renewcommand{\arraystretch}{1.35}
\begin{table*}[!t]
\centering
\begin{adjustbox}{width=\linewidth}
\begin{tabular}{@{}lccccccccccc@{}}
\toprule
Dataset                   & Train set?                & Dialogue Modality & Persona Modality      & Multple Session?          & Collection         & \# of E. & \# of S. & \# of I. & Avg. U./S. & Avg. I./E. & Avg. I./S. \\ \midrule
MMDD~\cite{lee2021constructing}                      & $\checkmark$ & T, V              & $\times$ & $\times$     & VSRN + Human       & -              & 17K            & 17K         & 12.74      & -          & 1.76       \\
PhotoChat~\cite{zang2021photochat}                 & $\checkmark$ & T, V              & $\times$ & $\times$     & Human              & -              & 11K            & 10K         & 11.56      & -          & 1          \\
MMDialog~\cite{feng2022mmdialog}                  & $\checkmark$ & T, V              & $\times$ & $\times$     & Social media       & -              & 1M             & 1.5M        & 4.56       & -          & 2.82       \\
DialogCC~\cite{lee2024dialogcc}                  & $\checkmark$ & T, V              & $\times$ & $\times$     & GPT-4, CLIP        & -              & 83K            & 120K        & 8.2        & -          & 7.83       \\
SODA~\cite{kim2022soda}                      & $\checkmark$ & T                 & $\times$ & $\times$     & InstructGPT        & -              & 1M             & -           & 7.6        & -          & -          \\
MSC~\cite{xu2021beyond} (train; 1-4 sessions) & $\checkmark$ & T                 & T                     & $\checkmark$ & Human              & 5K             & 16K            & -           & 13.4       & -          & -          \\
Conversation Chronicles~\cite{jang2023conversation}   & $\checkmark$ & T                 & T                     & $\checkmark$ & ChatGPT            & 200K           & 1M             & -           & 11.7       & -          & -          \\
\textsc{LoCoMo}~\cite{maharana2024evaluating}                    & $\times$     & T, V              & T                     & $\checkmark$ & ChatGPT + Human    & 50             & 1K             & 2K          & 15.8       & 32.3       & 3.72       \\
\grayrow \dataset \textbf{(Ours)}                     & $\checkmark$ & T, V              & T, V                  & $\checkmark$ & ChatGPT, Diffusion & 93K            & 0.5M           & 0.9M        & 10.5       & 9.94       & 1.86       \\ \bottomrule
\end{tabular}
\end{adjustbox}
\caption{Comparison of \dataset with existing datasets in terms of multi-modality and long-range continuity: MMDD, PhotoChat, MMDialog, DialogCC, SODA, MSC, Conversation Chronicles, and \textsc{LoCoMo}. V and T denote virtual and textual modality, respectively. E., S., and I. denote episode, session, and image, respectively. I.E. and I.S. denote images by episode and images by a single session, respectively. \dataset is the first to achieve a long-term multi-modal conversation that covers multi-modal persona information and includes a large scale, which leads to a well-generalized multi-modal conversation model. VSRN~\cite{li2019visual} is the text-image matching model.}
\label{main_tab:dataset_stat}
\vspace{-1em}
\end{table*}}

\section{Analysis of \datasetName} \label{sec:stark}

In this section, we conduct comprehensive analysis of \datasetName in terms of diverse perspectives: Comparison analysis to existing datasets (\cref{sec:comparison_analysis}), multifaceted analysis (\cref{sec:rich_info}), and human evaluation (\cref{sec:human_eval}).

\subsection{Comparison to Existing Datasets} \label{sec:comparison_analysis}

In Table~\ref{main_tab:dataset_stat}, we compare \datasetName with other existing datasets in terms of multi-modality and long-term continuity. In summary, \datasetName uniquely accomplishes a long-term multi-modal conversation, encompassing extensive multi-modal persona information and featuring a comparable data scale (0.5M sessions) to \textsc{Soda} (1M) and Conversation Chronicles (1M). Unlike other multi-modal dialogue datasets, which focus on singular sessions, \datasetName achieves a significantly larger scale of session dialogues and images. Additionally, \datasetName stands out among long-term dialogue datasets by exclusively covering multi-modal dialogue and persona information, including social persona attributes and pre-stored device images. While the \textsc{LoCoMo} dataset also addresses long-term multi-modal conversations, it lacks multi-modal persona information and is limited in scale (50 episodes), being designed mainly for evaluation benchmarks. Therefore, \datasetName is the first to offer a large-scale long-term multi-modal conversation dataset, enabling the development of a well-generalized multi-modal dialogue model.

{\renewcommand{\arraystretch}{1.35}
\begin{table}[!t]
\centering
\begin{adjustbox}{width=0.9\linewidth}
\begin{tabular}{@{}cclclc@{}}
\toprule
\multicolumn{4}{c}{Demographic} & \multicolumn{2}{c}{Persona} \\ \cmidrule(r){1-4} \cmidrule(r){5-6}
Age/Gender & Ratio & Country     & Ratio & Entity          & Ratio \\ \cmidrule(r){1-2} \cmidrule(r){3-4} \cmidrule(r){5-6}
50-60      & 14.12 & China       & 7.85  & animal           & 4.32  \\
20-30      & 13.68 & USA         & 7.79  & profession       & 4.18  \\
60-70      & 13.29 & UK          & 6.73  & school name      & 2.68  \\
40-50      & 13.19 & Russia      & 6.43  & book author      & 2.68  \\
80-90      & 12.88 & India       & 5.75  & music artist     & 2.55  \\
30-40      & 12.1  & Japan       & 5.72  & music instrument & 2.41  \\
70-80      & 10.96 & Brazil      & 5.64  & subject          & 2.36  \\
10-20      & 9.76  & Germany     & 5.6   & food             & 2.35  \\ \cmidrule(r){1-2}
Male       & 51.29 & Italy       & 5.41  & sport            & 2.35  \\
Female     & 48.71 & South Korea & 5.23  & season           & 2.34  \\ \bottomrule
\end{tabular}
\end{adjustbox}
\caption{The ratio (\%) of age groups and gender, along with the ratio of Top-10 persona entity categories and countries in \dataset.}
\label{main_tab:rich}
\vspace{-1em}
\end{table}}

\subsection{Multifaceted Analysis} \label{sec:rich_info}

\paragraph{Demographic.} As shown in Table~\ref{main_tab:rich}, our dataset exhibits a fairly balanced distribution across age, gender, and country. This suggests that our dataset is less likely to introduce biases during model training. Among the age groups, individuals aged 50 to 60 are the most represented. This indicates the potential applicability of our dataset in scenarios where an AI assistant needs to continuously care for older users, as highlighted in recent studies~\cite{bae2022building,bae2022keep}. The gender distribution is nearly equal, implying a lower possibility of gender bias problem.

\paragraph{Social Persona.} We derive the ratio of the Top-10 persona entity categories corresponding to the generated persona entity key from ChatGPT (in~\cref{sec:persona}). As shown in Table~\ref{main_tab:rich}, we observe that the categories of personas most commonly encountered in our everyday surroundings, such as animals and professions, are the most prevalent. The remaining categories are evenly distributed. This indicates that our dataset is well-balanced, providing a comprehensive understanding of various personas without bias towards any specific category.

\begin{figure}[t]
    \centering
    \includegraphics[width=\linewidth]{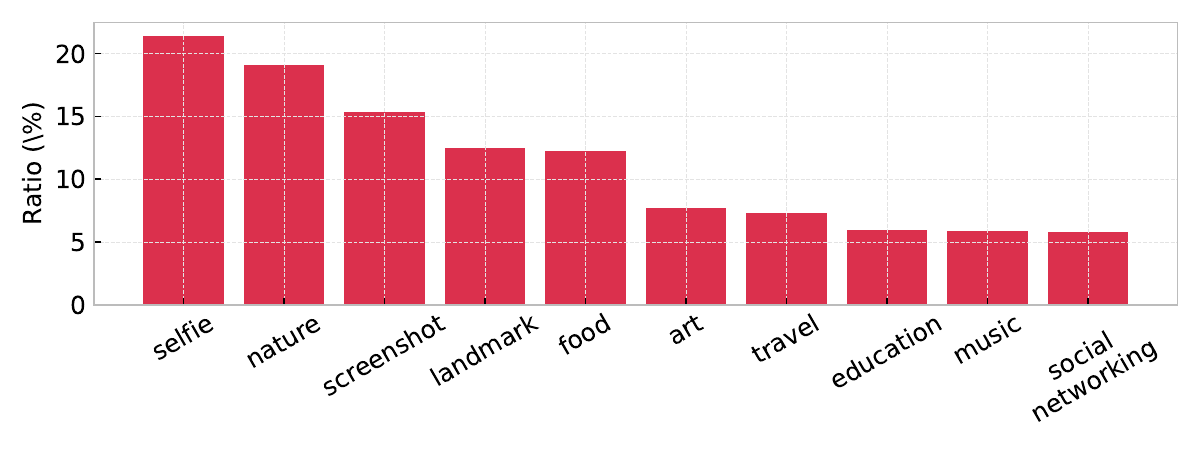}
    \caption{The ratio (\%) of Top-10 device image categories in \dataset.}
    \label{main_fig:device_category}
    \vspace{-1em}
\end{figure}

\begin{figure}[!t]
    \centering
    \includegraphics[width=0.9\columnwidth]{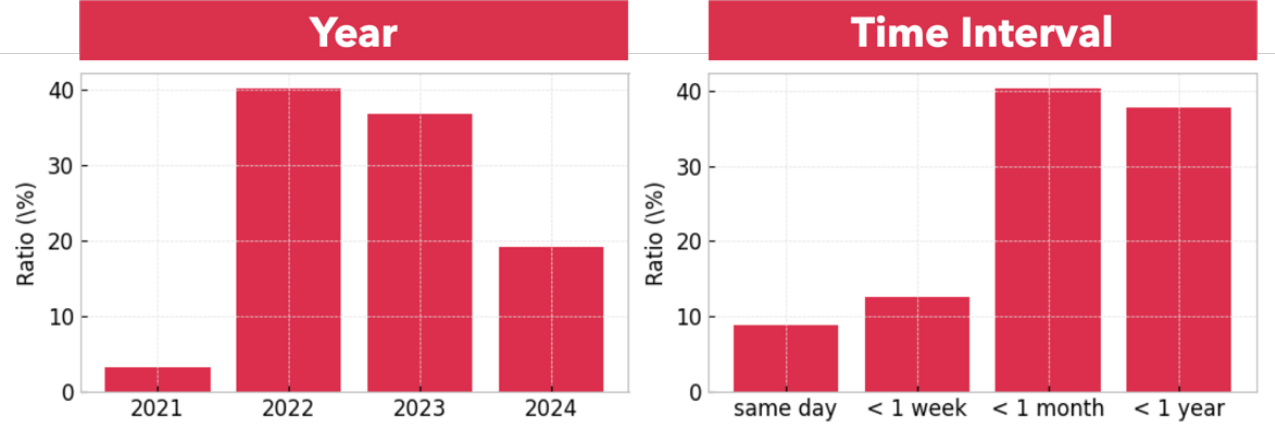}
    \caption{The distribution of year and time interval in \datasetName.}
    \label{main_fig:time_analysis}
    \vspace{-1em}
\end{figure}

\paragraph{Year \& Time Interval.} We analyze the distribution of year and time interval within \datasetName, as depicted in Figure~\ref{main_fig:time_analysis}. Our dataset predominantly contains conversations occurring from the year 2021 to 2024. The time intervals between sessions are frequently within one month, with a significant distribution of conversations occurring on the same day. This indicates that continuous care is realistically administered within short time intervals, demonstrating that our dataset effectively reflects real-world situations.

\paragraph{Device Image Category.}

Figure~\ref{main_fig:device_category} show the ratio of the Top-10 device image categories which are generated from ChatGPT (in \cref{sec:device}). We analyze a total of 467K device image categories. The prominent representation of categories such as selfies and nature landscape screenshots suggests that our dataset has a realistic distribution, reflecting what is commonly observed in real-world settings (\eg Instagram, Flickr platforms).

\subsection{Human Evaluation} \label{sec:human_eval}

To quantify the quality of \dataset, we conduct two different kinds of human evaluation, (1) human ratings and (2) head-to-head comparison, based on several evaluation criteria.

\paragraph{Human Ratings.} We meticulously evaluate the quality of \datasetName on seven distinct criteria: (1) coherence, (2) consistency, (3) image-sharing turn relevance, (4) image-dialogue relevance, (5) image-persona relevance, (6) time interval, and (7) experience. Each human evaluator rates 100 randomly chosen episode samples (totaling 500 session dialogues) using a detailed 4-point Likert scale for all criteria. Further explanations of each evaluation item and the recruitment process for human evaluators are provided in the Appendix~\ref{supp_sec:human_eval_question} and Appendix~\ref{supp_sec:detail_human_eval}. On average, we achieve significantly higher scores: 3.4 for coherence, 3.52 for consistency, 3.07 for image-sharing turn relevance, 2.49 for image-dialogue relevance, 3.35 for image-persona relevance, 3.75 for time interval, and 3.73 for experience. Additionally, we measure the inter-rater agreement (IA) using Krippendorff's $\alpha$, obtaining a value of 0.27, which indicates a fair level of agreement. These results underscore the reliability of \frameworkName in generating long-term multi-modal conversations starting with only basic demographic information.

\begin{figure}[t]
    \centering
    \includegraphics[width=\linewidth]{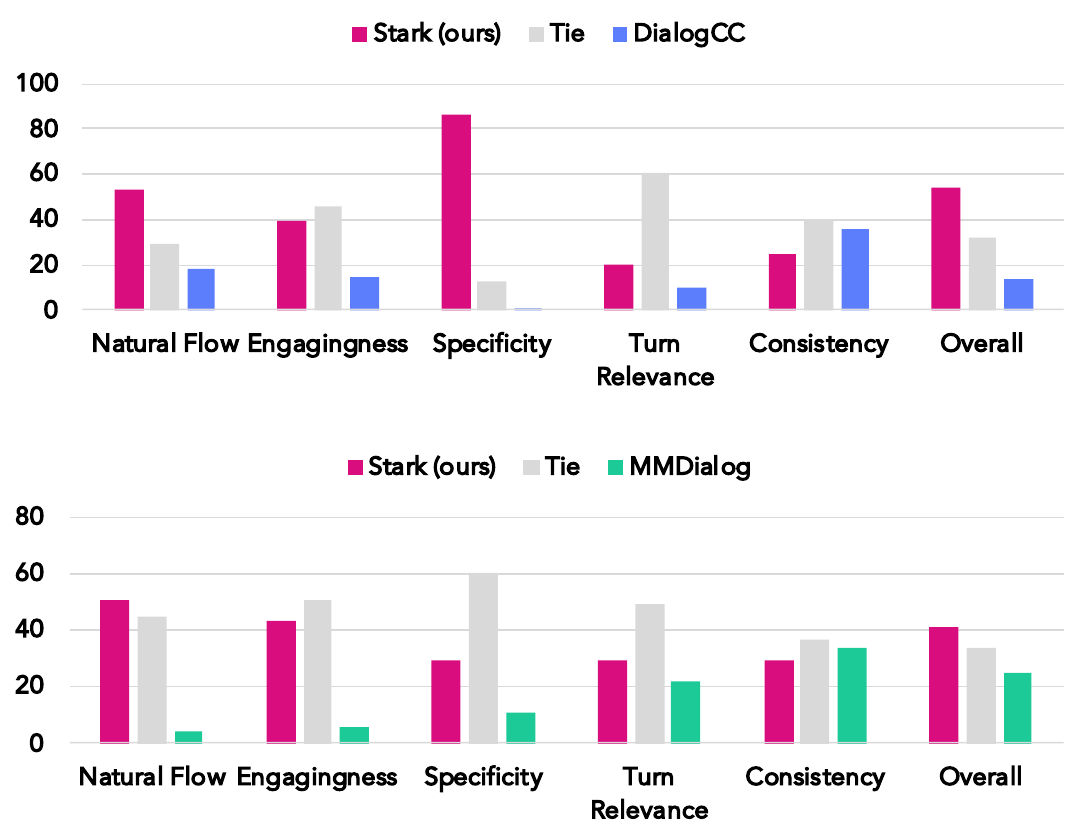}
    \caption{Results of head-to-head comparison between \dataset (ours) and two existing datasets, DialogCC~\cite{lee2024dialogcc} and MMDialog~\cite{feng2022mmdialog}, on six evaluation criteria.}
    \label{main_fig:head_to_head}
    \vspace{-1em}
\end{figure}

\paragraph{Head-to-Head Comparison.} Since \dataset is automatically constructed by leveraging various generative models, we assess the quality gap between our dataset and other high-quality and realistic datasets: DialogCC~\cite{lee2024dialogcc} (which has recently demonstrated high quality) and MMDialog~\cite{feng2022mmdialog} (which is derived from social media) by conducting a head-to-head comparison. Given that DialogCC and MMDialog are singular session datasets, we randomly sample 100 session dialogues from \datasetName and also randomly sample the same number of dialogues from DialogCC and MMDialog. We then evaluate them based on six criteria: (1) natural flow, (2) engagingness, (3) specificity, (4) image-sharing turn relevance, (5) image-dialogue consistency, and (6) overall quality. Further details are provided in the Appendix~\ref{supp_sec:human_eval_question}. Overall, as illustrated in Figure~\ref{main_fig:head_to_head}, \datasetName achieves better scores than both DialogCC and MMDialog across all criteria. Specifically, our dataset exhibits more engaging and naturally flowing conversations, particularly surpassing MMDialog by a large margin. Interestingly, human evaluators frequently select ``Tie'' for the items related to image-sharing turn relevance and image-dialogue consistency compared to other datasets. These results imply that, despite being constructed using generative models such as ChatGPT and our proposed image aligner (which includes several diffusion models), our dataset ensures the relevance of image-sharing moments and maintains the quality of generated images. This demonstrates the robustness and reliability of our proposed framework in producing coherent and engaging multi-modal conversations, even when compared to datasets utilizing actual photo-realistic images (\eg DialogCC uses CC3M~\cite{sharma2018conceptual}, MMDialog uses social media images).

\section{\model} \label{sec:ultron}

\begin{figure}
    \centering
    \includegraphics[width=0.8\linewidth]{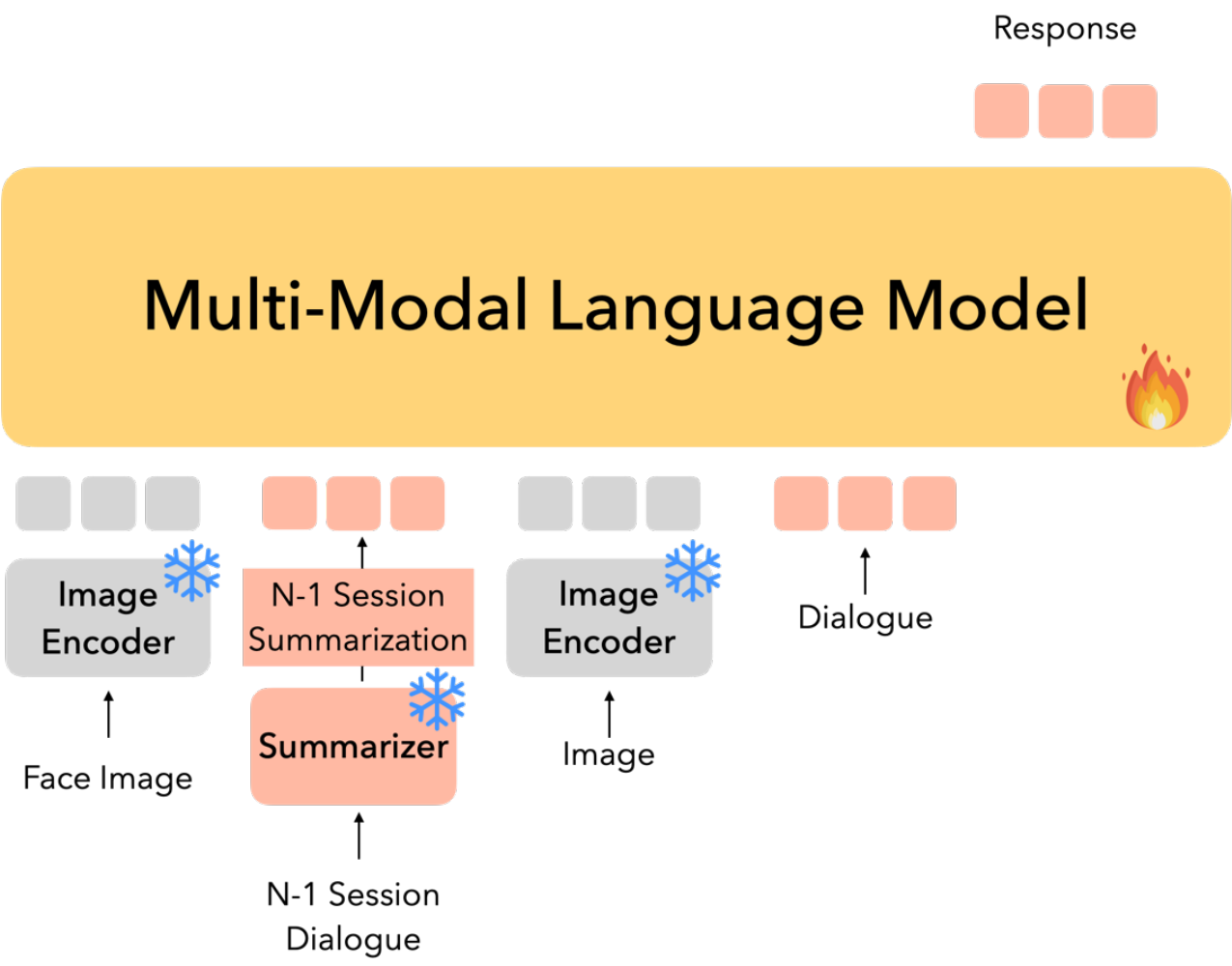}
    \caption{The overall architecture of \model.}
    \label{main_fig:ultron}
    \vspace{-1em}
\end{figure}

With \datasetName, we train a multi-modal conversation model named \model 7B. This model is designed to understand diverse social and personal dynamics along with previous interactions, enabling it to identify the appropriate moments for image sharing and retrieve relevant images based on the dialogue context. The overall architecture of \model is illustrated in Figure~\ref{main_fig:ultron}.

\subsection{Motivation behind Model Design}

\paragraph{Backbone Model.} Identifying the optimal moment for image sharing presents a significant challenge due to the subjective nature of this behavior, even for humans~\cite{lee2023large}. Additionally, retrieving relevant images based on dialogue context is non-trivial, as critical evidence is often dispersed throughout the entire conversation~\cite{chae2023dialogue,wang2023cue}. To address these challenges, we employ the recently proposed Meteor~\cite{lee2024meteor} model, which significantly enhances multi-modal reasoning capabilities across diverse tasks by introducing the novel concept of ``traversal of rationale.'' Consequently, we initiate the training of \model on the top of the Meteor model.

\paragraph{Input \& Output.} Recent studies have demonstrated the powerful visual imagination capabilities of large language models~\cite{lee2024dialogcc,lee2023large,li2024mini}. Inspired by these findings, we train \modelName to alternatively generate the image-sharing moment in a text format (without generating the image directly), specifically ``\texttt{<RET> <h> image description </h>}." This method allows \modelName to produce image descriptions that are better aligned with the given dialogue context, benefiting CLIP or generative models. In future work, since our model does not directly produce images, we will focus on developing a multi-modal language model capable of generating or retrieving images, following recent findings in the field~\cite{zheng2023minigpt,koh2024generating}.

\subsection{Model Architecture}

\modelName comprises a vision encoder, a vision projector, a summarizer, and the backbone multi-modal language model from the Meteor model. The architectures of the vision encoder, vision projector, and backbone model are consistent with those employed in the Meteor model. For the summarizer, we first construct a summarization dataset. Specifically, we randomly sample 10,000 episodes, encompassing a total of 53,317 session dialogues, and employ ChatGPT to generate summaries for these session dialogues. The prompt used for this task is detailed in the Appendix~\ref{supp_sec:prompt_template}. Utilizing this constructed dataset, we fine-tune the LLaMA-3 8B model~\cite{llama3modelcard}~\footnote{\url{https://huggingface.co/meta-llama/Meta-Llama-3-8B-Instruct}} with Q-LoRA~\cite{dettmers2024qlora} tuning, using 64 rank and 16 alpha parameters. This model is subsequently used to generate summaries for all session dialogues in our dataset. We then filter out unsuitable summaries, such as those containing repetition, ensuring that only high-quality summaries are included in the training dataset for \modelName.

\section{Experiments} \label{sec:expr}

\subsection{Experimental Setup} \label{sec:expr_setup}

\paragraph{Datasets.} To build generalized multi-modal conversation model that converse with user on diverse social situations, we train \modelName on \datasetName and Mini-Gemini Instruction~\cite{li2024mini}. We evaluate \modelName on PhotoChat~\cite{zang2021photochat}.

\vspace{-0.5em}

\paragraph{Task Definition.} We perform \modelName on dialogue-to-image retrieval task which is standard downstream task regarding multi-modal dialogue. The Dialogue-to-Image Retrieval task involves retrieving the relevant image based on the dialogue context.

\vspace{-0.5em}

\paragraph{Evaluation Metrics.} We use the widely adopted Recall@K and MRR metric. 

{\renewcommand{\arraystretch}{1.35}
\begin{table}[t]
\centering
\begin{adjustbox}{width=0.7\linewidth}

\begin{tabular}{@{}lcccc@{}}
\toprule
Model                 & R@1                  & R@5                  & R@10                 & MRR                  \\ \midrule
\grayrow \multicolumn{5}{l}{\textit{Fine-tuned Performance}}                                                                         \\
BM25                  & 6.6                  & 15.4                 & 23.0                 & -                    \\
DE                    & 9.0                  & 26.4                 & 35.7                 & -                    \\
VSE++                 & 10.2                 & 25.4                 & 34.2                 & -                    \\
SCAN                  & 10.4                 & 27.0                 & 37.1                 & -                    \\
VLMo                  & 13.8                 & 30.0                 & 39.4                 & -                    \\
ViLT                  & 11.5                 & 25.6                 & 33.8                 & -                    \\
PaCE                  & 15.2                 & 36.7                 & 49.6                 & -                    \\
DialCLIP & 19.5 & 44.0 & 55.8 & - \\
\grayrow \multicolumn{5}{l}{\textit{VLM, zero-shot}}                                                                                 \\
CLIP-base             & 13.7                 & 28.0                 & 35.2                 & 20.8                 \\
CLIP-large            & 14.1                 & 28.7                 & 35.3                 & 21.5                 \\
\grayrow \multicolumn{5}{l}{\textit{Large Multi-Modal Model}}                                                       \\
LLaVA v1.5 7B         & 11.1                 & 26.5                 & 33.3                 & 18.8                 \\
LLaVA v1.5 13B        & 12.1                 & 25.6                 & 32.3                 & 19.3                 \\
MiniGPT-4$_{\text{Vicuna 7B}}$  & 11.6                 & 26.5                 & 34.0                 & 19.1                 \\
MiniGPT-4$_{\text{Vicuna 13B}}$ & 11.7                 & 27.7                 & 35.5                 & 19.8                 \\
Qwen-VL-Chat 7B	& 12.1	& 27.4	& 36.1	& 20.2 \\
GPT4-V                & 13.8                 & 27.9                 & 35.9                 & 21.3                 \\
\grayrow \multicolumn{5}{l}{\textit{LLM-based Framework}}                                                                                         \\
\driber$_{\text{ChatGPT 0613}}$          & 26.6                 & 46.1                 & 54.2                 & 36.0                 \\
\driber$_{\text{ChatGPT 1106}}$          & 26.3                 & 45.6                 & 54.3                 & 35.4                 \\
\driber$_{\text{GPT-4 1106}}$           & 28.3                 & 47.4                 & 55.2                 & 37.6                 \\
\driber$_{\text{Vicuna-13B}}$            & 25.8                 & 45.0                 & 53.1                 & 35.0                 \\
\driber$_{\text{LLaMa2-Chat-70B}}$       & 24.5                 & 43.5                 & 52.6                 & 34.0                 \\ \midrule
\grayrow \model & \textbf{31.2} & \textbf{53.7} & \textbf{65.0} & \textbf{46.1} \\
\bottomrule

\end{tabular}

\end{adjustbox}
\caption{Comparison results of the dialogue-to-image retrieval task on PhotoChat~\cite{zang2021photochat}.}
\label{main_tab:retrieval_expr}

\end{table}}

\subsection{Results} \label{sec:expr_result}

As shown in Table~\ref{main_tab:retrieval_expr}, \modelName achieves significant performance improvements in the dialogue-to-image retrieval task compared to several other methods. Notably, \modelName outperforms the recent LLM-based framework, \driber~\cite{lee2023large}. Interestingly, recent large multi-modal models, such as LLaVA v1.5~\cite{liu2024visual} and GPT-4V~\cite{gpt4v}, exhibit relatively lower performance in a zero-shot setting. In contrast, \modelName achieves remarkable performance, underscoring the effectiveness of our dataset in enhancing complex image-sharing behaviors.

\section{Conclusion}

In this work, we first propose a social long-term multi-modal conversation dataset, \dataset, which is fully automatically constructed through our proposed framework, \frameworkName. This framework comprises ChatGPT and our proposed \planExecute image aligner. Through extensive experiments, we demonstrate that our dataset has comparable quality to other existing datasets. Additionally, using our dataset, we build a multi-modal conversation model, \model, which achieves significant performance in the dialogue-to-image retrieval task.
\section*{Limitations}

\paragraph{Inconsistent Personalized Images.} 

To construct a dataset encompassing personalized image-sharing behavior, we utilized a personalized text-to-image generative model. However, this occasionally led to instances where the appearance of the user was not consistently maintained across some samples. Additionally, when generating images featuring groups, there was a tendency for multiple individuals in the group to appear identical to the user's appearance. Despite applying various filtering methods to mitigate these issues, complete elimination was not achieved. Given the rapid advancements in generative models, we anticipate that future, more advanced models will enable the creation of datasets with enhanced consistency.

\paragraph{Building Role-Specified AI Assistant.}

When constructing our dataset, we did not provide the AI assistant with any specific personality traits or preference information~\cite{lee2024aligning}. For future research, it would be advantageous to develop datasets or models that incorporate social relational information~\cite{zhou2023sotopia,jang2023conversation} (\eg friend, colleague), a broader range of conversational styles~\cite{han2022meet}, and personality traits. This approach could enhance social interactions and foster a closer relationship between the AI assistant and users.

\section*{Ethical Considerations}

Despite applying various filtering methods to exclude unsuitable samples, potential issues may still exist within our proposed framework.
Firstly, the generated dialogue might propagate social or cultural biases, as ChatGPT can produce harmful content, including social biases and offensive remarks \citep{baheti2021just,hartvigsen2022toxigen}.
Secondly, the generated images may also reflect unfaithful and socially biased content when using Stable Diffusion \citep{rombach2022high}. As reported by \citep{wang2021gender}, even when providing gender-neutral queries to the CLIP model \citep{radford2021learning}, the model occasionally retrieves images that cause gender-bias issues.
We are concerned that these problematic issues may persist in the augmented dataset. Consequently, a multi-modal dialogue model trained on this dataset might sometimes generate or retrieve biased images. It is crucial to consider these issues carefully when developing a multi-modal dialogue model.

\section*{Acknowledgement}

This work was supported by a grant of the KAIST-KT joint research project through AI Tech Lab, Institute of convergence Technology, funded by KT [Project No. G01230605, Development of Task-oriented Persona-based Dialogue Generation Combining Multi-modal Interaction and Knowledge Modeling].

\bibliography{custom}

\appendix

\section{Detailed Explanation of \frameworkName} \label{supp_sec:detail_framework}

\subsection{Pre-defined Demographic Lexicon} \label{supp_sec:demographic_lexicon}

\paragraph{Age Group.} The age groups are defined as follows: \texttt{10-20}, \texttt{20-30}, \texttt{30-40}, \texttt{40-50}, \texttt{50-60}, \texttt{60-70}, \texttt{70-80}, and \texttt{80-90}. From these groups, a group is first selected at random. Subsequently, an age within the selected group is chosen randomly. For example, if the \texttt{10-20} group is selected, a number between 10 and 20 is then randomly chosen.

\paragraph{Gender.} We consider two gender categories: \texttt{male} and \texttt{female}, with the selection made randomly. Although it is essential to consider fairness, including non-binary gender categories for fair AI practices, the current attribute lexicon for human face generation does not support non-binary options. Therefore, we have excluded it to maintain the quality of the generated human face images. In future work, we aim to incorporate socially-aware fairness in our \frameworkName to develop a socially-balanced multi-modal dialogue dataset.

\paragraph{Birthplace \& Residence.} To determine the birthplace and residence, we first prepare a country list, as referenced from previous work~\cite{santy2023nlpositionality}, that includes 19 countries: \texttt{United States of America}, \texttt{China}, \texttt{Japan}, \texttt{India}, \texttt{United Arab Emirates}, \texttt{France}, \texttt{Germany}, \texttt{Italy}, \texttt{South Korea}, \texttt{Saudi Arabia}, \texttt{Kazakhstan}, \texttt{Brazil}, \texttt{Mexico}, \texttt{Egypt}, \texttt{Argentina}, \texttt{Russia}, \texttt{United Kingdom}, \texttt{Spain}, and \texttt{Canada}. We randomly select a country from this list to assign as the birthplace and residence. In 70\% of the cases, the birthplace and residence are the same, while in 30\% of the cases, the birthplace and residence are different (\eg due to immigration).

{\renewcommand{\arraystretch}{1.35}
\begin{table*}[!t]
\centering
\begin{adjustbox}{width=\linewidth}
\begin{tabular}{@{}ll@{}}
\toprule
Persona Commonsense Relation       & Template for Sentence Form \\ \midrule
Routines/Habits  &  My name is \textcolor{blue}{\texttt{\{name\}}}. \textcolor{blue}{\texttt{\{demographic sentence\}}} \textcolor{blue}{\texttt{\{persona sentence\}}} I regularly \textcolor{blue}{\texttt{\{commonsense\}}}. \\
Characteristics &      My name is \textcolor{blue}{\texttt{\{name\}}}. \textcolor{blue}{\texttt{\{demographic sentence\}}} \textcolor{blue}{\texttt{\{persona sentence\}}} I \textcolor{blue}{\texttt{\{commonsense\}}}.                      \\
Experiences     &     My name is \textcolor{blue}{\texttt{\{name\}}}. I \textcolor{blue}{\texttt{\{commonsense\}}}. Now, \textcolor{blue}{\texttt{\{demographic sentence\}}} \textcolor{blue}{\texttt{\{persona sentence\}}}                       \\
Goals/Plans      &         My name is \textcolor{blue}{\texttt{\{name\}}}. \textcolor{blue}{\texttt{\{demographic sentence\}}} \textcolor{blue}{\texttt{\{persona sentence\}}} I plan \textcolor{blue}{\texttt{\{commonsense\}}}.                   \\
Relationships   &       My name is \textcolor{blue}{\texttt{\{name\}}}. \textcolor{blue}{\texttt{\{demographic sentence\}}} \textcolor{blue}{\texttt{\{persona sentence\}}} So, I \textcolor{blue}{\texttt{\{commonsense\}}}.                     \\ \bottomrule
\end{tabular}
\end{adjustbox}
\caption{Templates (used in \cref{sec:narrative}) for converting persona-related commonsense knowledge represented in a symbolic format to short sentence form. \textcolor{blue}{\texttt{\{demographic sentence\}}} also consists of short template which is represented in the format: I am a \textcolor{blue}{\texttt{\{age\}}}-year-old \textcolor{blue}{\texttt{\{gender\}}}. I was born in \textcolor{blue}{\texttt{\{birthplace\}}}, I currently reside in \textcolor{blue}{\texttt{\{residence\}}}. \textcolor{blue}{\texttt{\{persona sentence\}}} is generated from ChatGPT (\cref{sec:persona}) and \textcolor{blue}{\texttt{\{commonsense\}}} is generated from ChatGPT (\cref{sec:commonsense}).} 
\label{supp_tab:sentence_form}
\vspace{-1em}
\end{table*}}

\subsection{Social Persona Categories} \label{supp_sec:persona_category}

{\renewcommand{\arraystretch}{1.35}
\begin{table}[!t]
\centering
\begin{adjustbox}{width=\linewidth}
\begin{tabular}{@{}lccccc@{}}
\toprule
Step Name        & Temp. & Top-p & Freq. & Pres. & Max tokens \\ \midrule
Persona (\cref{sec:persona})          & 0.9         & 1     & 0    & 0    & 2048       \\
Persona Commonsense (\cref{sec:commonsense})     & 0.9         & 1     & 0    & 0    & 1024       \\
Personal Narrative (\cref{sec:narrative})      & 0.9         & 0.95  & 1    & 0.6  & 2048       \\
Event Sequence (\cref{sec:event})           & 0.9         & 1     & 0    & 0    & 4096       \\
Device (\cref{sec:device})           & 0.9         & 1     & 0    & 0    & 1024       \\
Dialogue (\cref{sec:conversation})         & 0.9         & 0     & 0    & 0    & 4096       \\
\planExecute (\cref{sec:planExecute}) & 0.9         & 0.95  & 1    & 0.6  & 1024       \\ \cdashline{1-6}\noalign{\vskip 0.5ex}
Dialogue Summary (\cref{sec:ultron}) & 0.9         & 0.95  & 1    & 0.6  & 1024       \\
\bottomrule
\end{tabular}
\end{adjustbox}
\caption{ChatGPT generation settings, including temperature (Temp.), top-p (Top-p), frequency penalty (Freq.), presence penalty (Pres.), and the maximum number of tokens (Max tokens) configured for each step within our proposed \frameworkName framework and for dialogue summarization (\cref{sec:ultron}).} 
\label{supp_tab:implementation_framework}
\vspace{-1em}
\end{table}}

{\renewcommand{\arraystretch}{1.35}
\begin{table}[!t]
\centering
\begin{adjustbox}{width=\linewidth}
\begin{tabular}{@{}ll@{}}
\toprule
Persona Category                            & Entity Key          \\ \midrule
School $\supset$ Name                               & school name         \\
School $\supset$ Type                               & school type         \\
Employment $\supset$ Company                        & company name        \\
Employment $\supset$ Workplace                      & workplace           \\
School $\supset$ Degree                             & degree              \\
School $\supset$ Degree Subject                     & degree subject      \\
Employment $\supset$ Profession                     & profession          \\
Possession $\supset$ Animal                         & animal              \\
Possession $\supset$ Vehicle                        & vehicle             \\
Employment $\supset$ Job Status                     & job status          \\
Preference $\supset$ Location                       & location            \\
Preference $\supset$ Place                          & place               \\
Preference $\supset$ Show                           & show                \\
Preference $\supset$ Media Genre                    & media genre         \\
Preference $\supset$ Animal                         & animal              \\
Preference $\supset$ Book Author                    & book author         \\
Preference $\supset$ Book Genre                     & book genre          \\
Preference $\supset$ Book Title                     & book title          \\
Preference $\supset$ Color                          & color               \\
Preference $\supset$ Drink                          & drink               \\
Preference $\supset$ Food                           & food                \\
Preference $\supset$ Hobby                          & hobby               \\
Preference $\supset$ Movie Genre                    & movie genre         \\
Preference $\supset$ Movie Title                    & movie\_title        \\
Preference $\supset$ Music Genre                    & music genre         \\
Preference $\supset$ Music Instrument               & music instrument    \\
Preference $\supset$ Music Artist                   & music artist        \\
Preference $\supset$ Season                         & season              \\
Preference $\supset$ Sport                          & sport               \\
Location $\supset$ Residence                        & city-state          \\
Location $\supset$ Residence                        & country             \\
Employment $\supset$ Previous Profession            & profession          \\
Employment $\supset$ Teaching Experience $\supset$ Subject  & subject             \\
Employment $\supset$ Teaching Experience $\supset$ Activity & activity            \\
School $\supset$ Status                             & school status       \\
Physical Symptom                            & physical symptom    \\
Psychiatric Symptom                         & psychiatric symptom \\
Respiratory Disease                         & respiratory disease \\
Digestive Disease                           & digestive disease   \\
Medicine                                    & medicine            \\
Preference $\supset$ Game                           & game                \\
Preference $\supset$ Fashion                        & fashion             \\
Preference $\supset$ Social Media                   & social media        \\
Preference $\supset$ Health \& Fitness              & health \& fitness   \\
Preference $\supset$ Technology                     & technology          \\
Preference $\supset$ Art \& Design                  & art \& design       \\
Preference $\supset$ Travel                         & travel              \\
Preference $\supset$ Politic                        & politic             \\
Preference $\supset$ Social Issue                   & social issue        \\
Preference $\supset$ Science                        & science             \\ \bottomrule

\end{tabular}
\end{adjustbox}
\caption{Social persona categories with corresponding persona entity keys. The symbol ``$\supset$'' indicates inclusion, representing a hierarchical category structure.} 
\label{supp_tab:persona_category}
\vspace{-1em}
\end{table}}

Table~\ref{supp_tab:persona_category} lists all the persona categories along with their corresponding persona entity keys. These categories are utilized in generating social persona sentences, as described in \cref{sec:persona}.

\subsection{Human Attribute Pool} \label{supp_sec:human_attr_pool}

To generate a virtual human face for consistent personalized image-sharing behavior, we leverage predefined human attribute information introduced by a recent work~\cite{li2024cosmicman}. In total, we use 23 human attributes: \texttt{style}, \texttt{body shape}, \texttt{background}, \texttt{hair}, \texttt{special clothing}, \texttt{one-piece outfits}, \texttt{tops}, \texttt{coats}, \texttt{bottoms}, \texttt{shoes}, \texttt{bags}, \texttt{hats}, \texttt{belts}, \texttt{scarves}, \texttt{headbands}, \texttt{headscarves}, \texttt{veils}, \texttt{socks}, \texttt{ties}, \texttt{age}, \texttt{gender}, and \texttt{birthplace}. In the predefined human attribute pool, each sample consists of a combination of these attributes. We randomly sample one combination from the human attribute pool. For example, ``\textit{A upper body shot, a 42-years-old female from Japan, fit, a white wall, wavy black below chest hair, red high neck normal long sleeve cotton solid color sweaters, red close-fitting maxi cotton plaid pants, black ankle boots leather solid color high heels, cotton solid color socks, cotton plaid tie.}''.

\subsection{Implementation Details}

Table~\ref{supp_tab:implementation_framework} present the generation settings of ChatGPT, which are used in our \frameworkName framework. 

\section{Human Evaluation Questionnaire} \label{supp_sec:human_eval_question}

This section presents the list of questions and multiple-choice options used for two human evaluations reported in Section 3.4: human ratings and head-to-head comparison. 

\subsection{Human Ratings} \label{supp_sec:human_rating}

\begin{itemize}
    \item \textbf{Coherence:} Do you think the conversation between the two speakers (\ie user, AI assistant) has a natural flow regarding event transitions?
    \begin{description}
        \item [Options:] 1: Not at all / 2: A little / 3: Somewhat / 4: A lot
    \end{description}
    
    \item \textbf{Consistency:} Do you think two speakers (\ie user, AI assistant) do not make a contradiction from past sessions?
    \begin{description}
        \item [Options:] 1: Not at all / 2: A little / 3: Somewhat / 4: A lot
    \end{description}
    
    \item \textbf{Image-Sharing Turn Relevance:} Do you think the image-sharing turn in the given dialogue is appropriate?
    \begin{description}
        \item [Options:] 1: Not at all / 2: A little / 3: Somewhat / 4: A lot
    \end{description}
    
    \item \textbf{Image-Dialogue Relevance:} How relevant do you think the aligned image is based on the dialogue context? 
    \begin{description}
        \item [Options:] 1: Not at all / 2: A little / 3: Somewhat / 4: A lot
    \end{description}

    \item \textbf{Image-Persona Relevance:} Does the aligned image accurately reflect the user's characteristics?
    \begin{description}
        \item [Options:] 1: Not at all / 2: A little / 3: Somewhat / 4: A lot
    \end{description}
    
    \item \textbf{Time Interval:} Do two speakers (\ie user, AI assistant) appear conversing in each session as though the designated time has passed since the previous session?
    \begin{description}
        \item [Options:] 1: Not at all / 2: A little / 3: Somewhat / 4: A lot
    \end{description}

    \item \textbf{Experience:} Do you think the user's experience is well reflected in the current session?
    \begin{description}
        \item [Options:] 1: Not at all / 2: A little / 3: Somewhat / 4: A lot
    \end{description}
    
\end{itemize}

\subsection{Head-to-Head Comparison} \label{supp_sec:head_to_head}

\begin{itemize}
    \item \textbf{Natural Flow:} Which dialogue has a more natural flow?
    \begin{description}
        \item [Options:] A / Tie / B
    \end{description}
    
    \item \textbf{Engagingness:} Which dialogue is more interesting and engaging?
    \begin{description}
        \item [Options:] A / Tie / B
    \end{description}

    \item \textbf{Specificity:} Which dialogue is more specific?
    \begin{description}
        \item [Options:] A / Tie / B
    \end{description}
    
    \item \textbf{Image-Sharing Turn Relevance:} Which dialogue has a more appropriate image-sharing turn?
    \begin{description}
        \item [Options:] A / Tie / B
    \end{description}
    
    \item \textbf{Image-Dialogue Consistency:} Which dialogue is more consistent between aligned image and dialogue context? 
    \begin{description}
        \item [Options:] A / Tie / B
    \end{description}
    
    \item \textbf{Overall:} Which dialogue has higher quality overall?
    \begin{description}
        \item [Options:] A / Tie / B
    \end{description}
    
\end{itemize}

\section{Human Evaluation System}

\begin{figure}
    \centering
    \includegraphics[width=\linewidth]{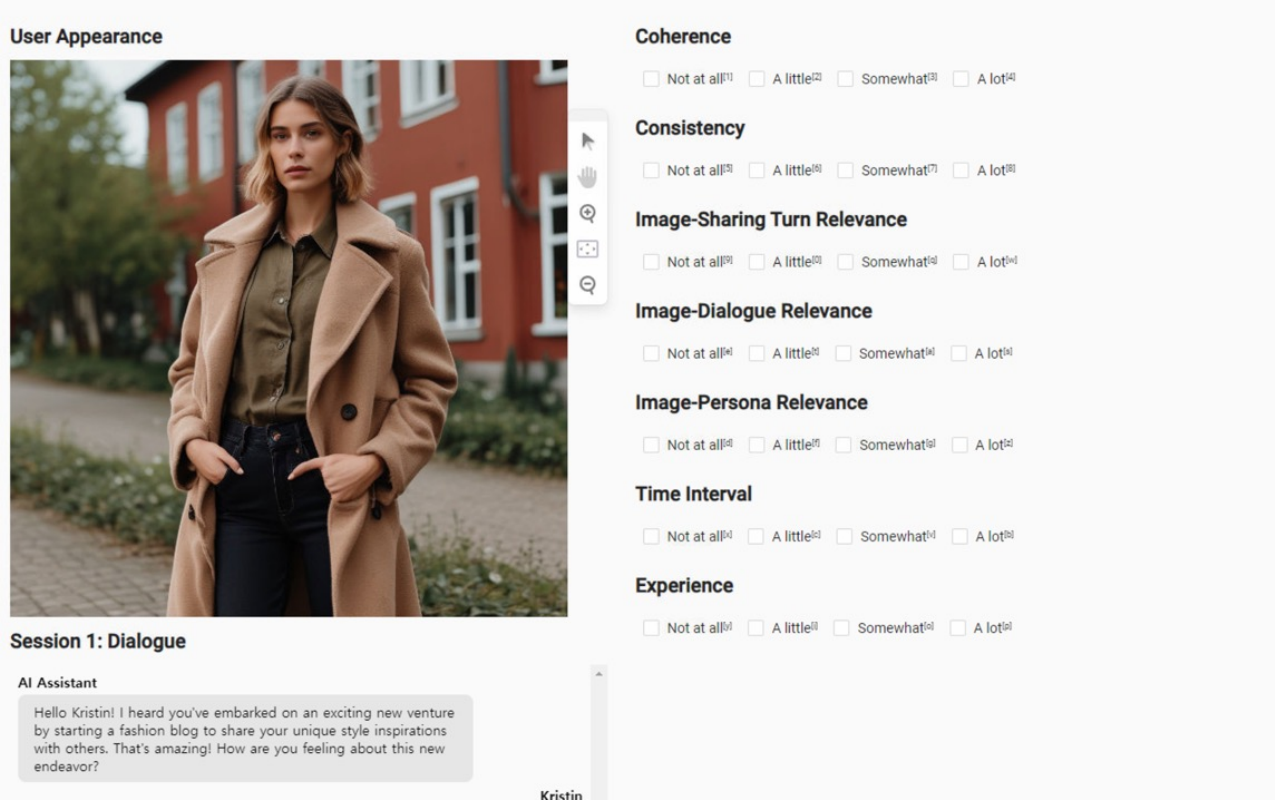}
    \caption{A screenshot of human rating evaluation.}
    \label{supp_fig:rating_system}
\end{figure}

\begin{figure}
    \centering
    \includegraphics[width=\linewidth]{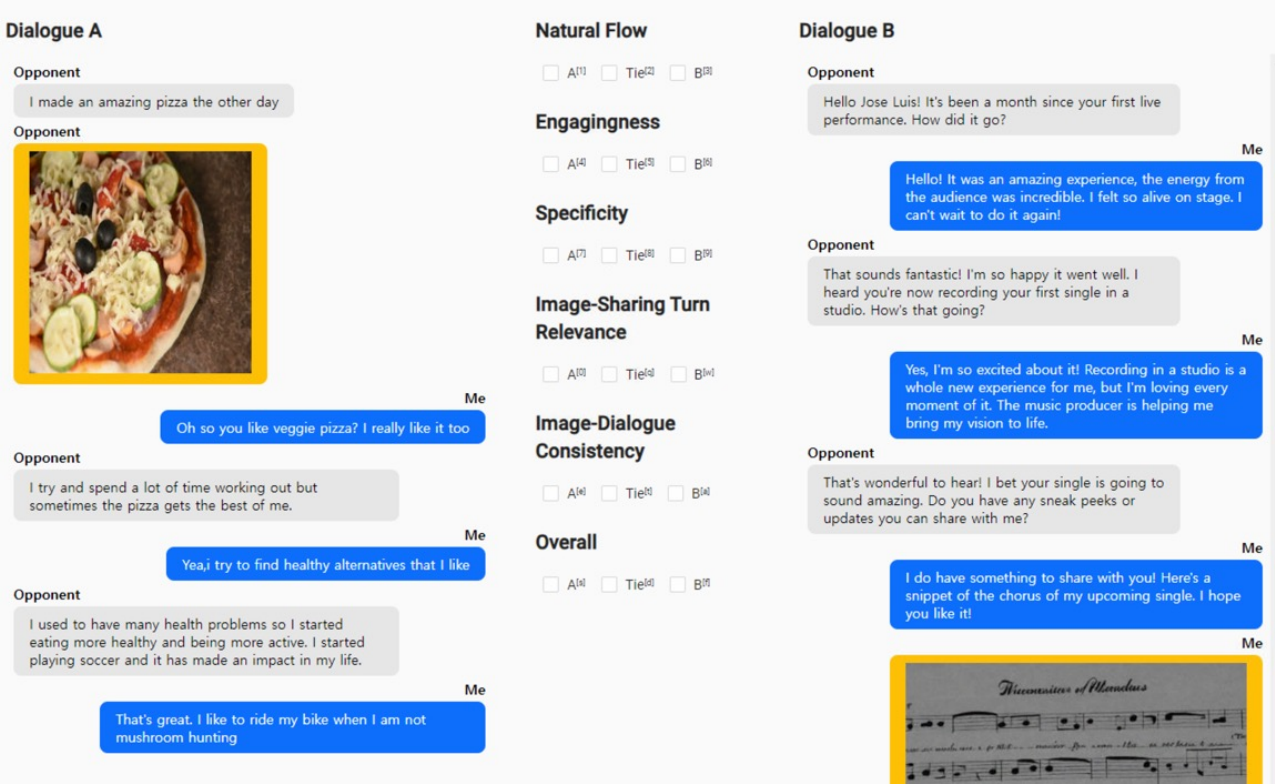}
    \caption{A screenshot of head-to-head comparison evaluation.}
    \label{supp_fig:ab_system}
\end{figure}

We show a screenshot of the human evaluation system in Figure~\ref{supp_fig:rating_system} and Figure~\ref{supp_fig:ab_system}. We implement this system using Label Studio \cite{LabelStudio}.

\section{Details of Human Evaluation} \label{supp_sec:detail_human_eval}
We recruited 9 individuals, unknown to us, who are either graduate or undergraduate students. Prior to participating in the experiment, they were provided with comprehensive instruction on the task, an overview of the multi-modal dialogue dataset, and a detailed explanation of the evaluation criteria. This preparatory phase lasted approximately one hour.

\onecolumn

\section{A Full Example of \dataset} \label{supp_sec:full_example}

\begin{figure}[ht]
    \centering
    \includegraphics[width=\linewidth]{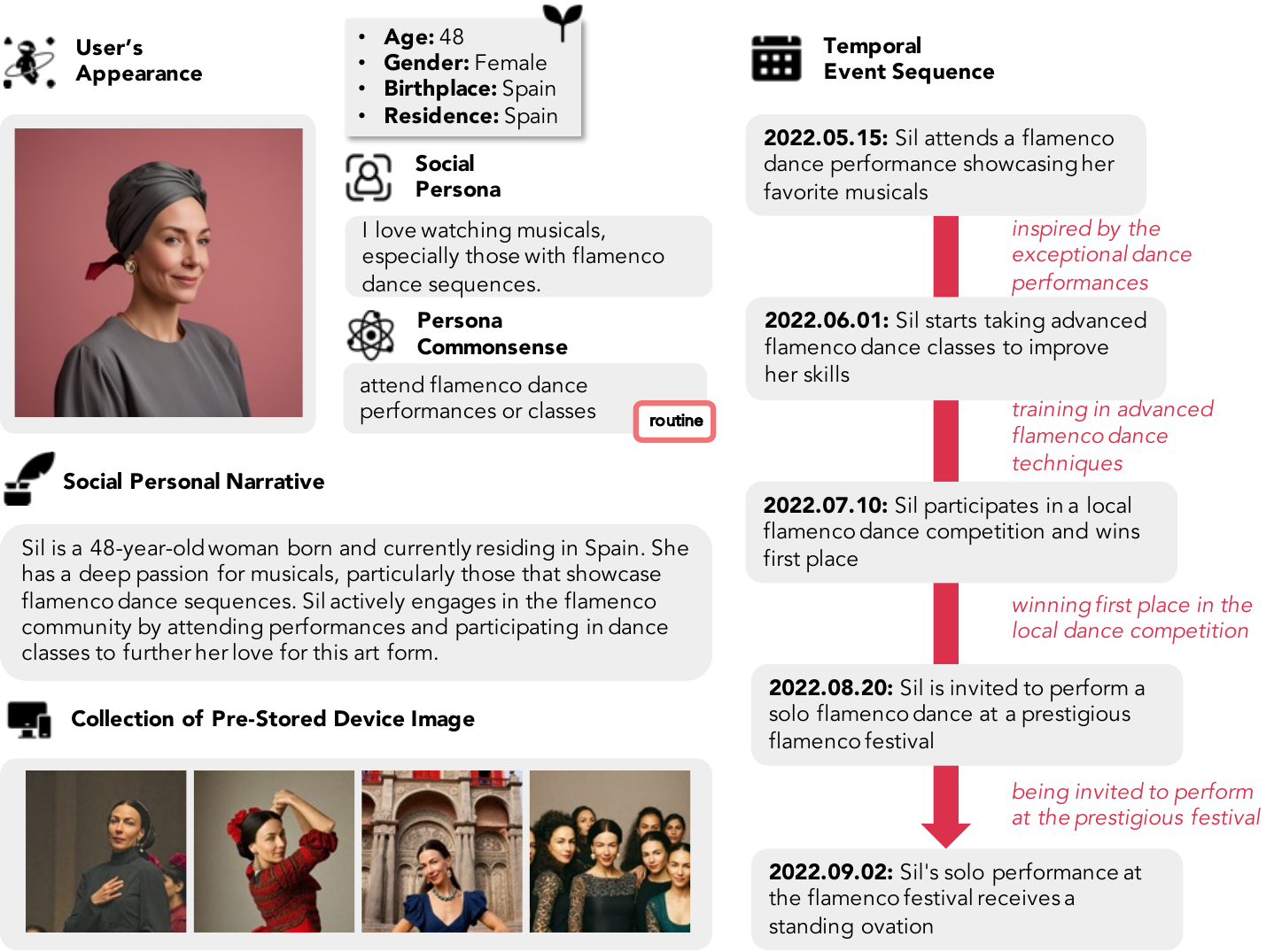}
    \caption{An example of \dataset in terms of demographic information, user's appearance, social persona, persona commonsense knowledge (\ie routine relation), personal narrative, collection of pre-stored device images, and temporal event sequences.}
    \label{supp_fig:case_study_1_1}
\end{figure}

\begin{figure}[ht]
    \centering
    \includegraphics[width=\linewidth]{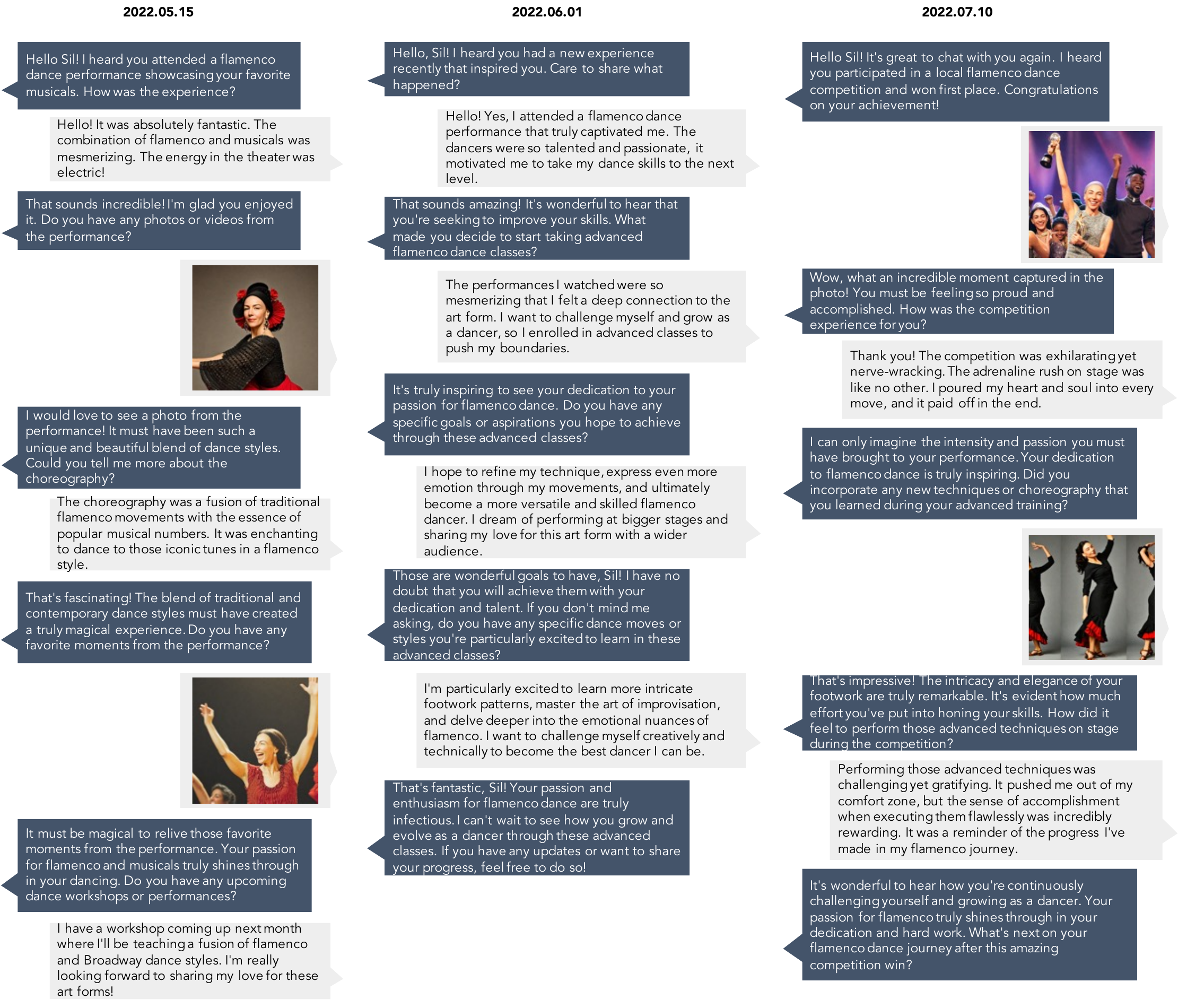}
    \caption{An example of \dataset regarding the temporal event sequences, as presented in Figure~\ref{supp_fig:case_study_1_1}. The left side shows the responses from the AI assistant, while the right side shows the responses from the user.}
    \label{supp_fig:case_study_1_2}
\end{figure}

\begin{figure}[ht]
    \centering
    \includegraphics[width=\linewidth]{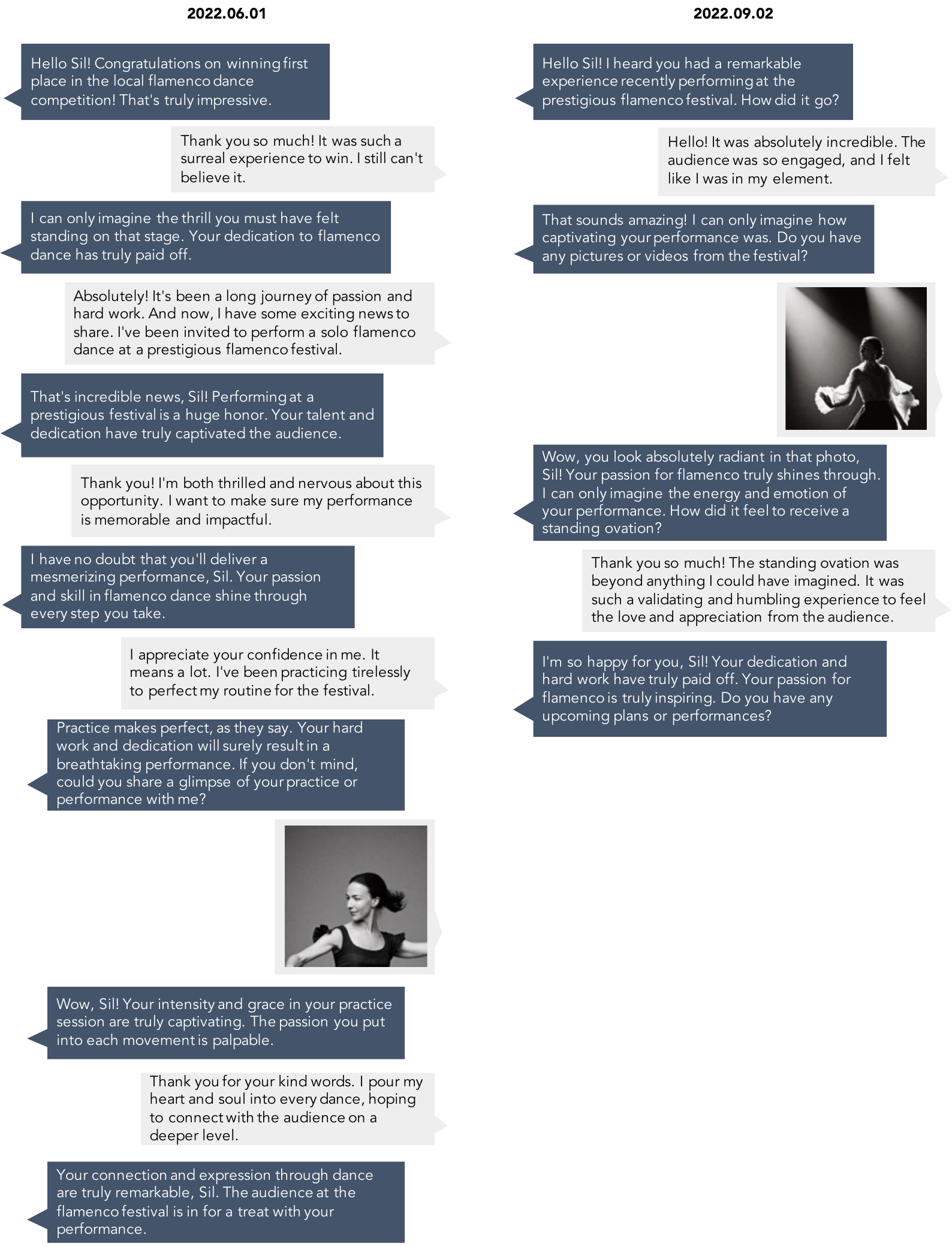}
    \caption{An example of \dataset regarding the temporal event sequences, as presented in Figure~\ref{supp_fig:case_study_1_1}. The left side shows the responses from the AI assistant, while the right side shows the responses from the user.}
    \label{supp_fig:case_study_1_3}
\end{figure}

In this section, we show a full conversation of \dataset in Figure~\ref{supp_fig:case_study_1_1}, Figure~\ref{supp_fig:case_study_1_2}, Figure~\ref{supp_fig:case_study_1_3}.

\clearpage

\section{Prompt Templates}\label{supp_sec:prompt_template}

\begin{prompt}{Prompt Template for Social Persona Attribute Generation}
    \textbf{{\system}:}\\
    \\
    Based on the given persona category, entity key, and user's profile information (i.e., age, gender, nationality), your job is to generate 30 persona sentences and corresponding persona entity values in the format "<persona sentence> (<entity key>: <entity value>)." You should generate very specific persona sentences and entity values. The persona sentence can express a positive sentiment (like) or a negative one (dislike). \\
    \\
    For example, \\
    \\    
    \textcolor{blue}{\texttt{\{few-shot example\}}}\\
    \tcblower
    
    \textbf{{\user}:}\\
    \\
    Profile Information:\\
    - Age: \textcolor{blue}{\texttt{\{age\}}}\\
    - Gender: \textcolor{blue}{\texttt{\{gender\}}}\\
    - Birthplace: \textcolor{blue}{\texttt{\{birthplace\}}}\\
    - Residence: \textcolor{blue}{\texttt{\{residence\}}}\\
    \\
    Persona Category: \textcolor{blue}{\texttt{\{target persona category\}}}\\
    Persona Entity Key: \textcolor{blue}{\texttt{\{target persona entity\}}}\\
    Persona Sentences:\\
    1.
\end{prompt}

\begin{prompt}{Prompt Template for Social Persona Commonsense Generation: \Routine}
    \textbf{{\system}:}\\
    \\
    You are a helpful assistant. \\
    \tcblower
    
    \textbf{{\userRoutine}:}\\
    \\
    \textcolor{blue}{\texttt{\{demographic sentence\}}} \textcolor{blue}{\texttt{\{persona sentence\}}} I regularly <routine/habit>.\\
    \\    
    Generate the most appropriate sentence for "<routine/habit>" in the given sentence. You must provide the answer corresponding to "<routine/habit>".\\
    <routine/habit>:
\end{prompt}

\begin{prompt}{Prompt Template for Social Persona Commonsense Generation: \Goal}
    \textbf{{\system}:}\\
    \\
    You are a helpful assistant. \\
    \tcblower
    
    \textbf{{\userGoal}:}\\
    \\
    \textcolor{blue}{\texttt{\{demographic sentence\}}} \textcolor{blue}{\texttt{\{persona sentence\}}} I plan <goal/plan>. \\
    \\
    Generate the most appropriate sentence for "<goal/plan>" in the given sentence. You must provide the answer corresponding to "<goal/plan>".\\
    <goal/plan>: 
\end{prompt}

\begin{prompt}{Prompt Template for Social Persona Commonsense Generation: \Relationship}
    \textbf{{\system}:}\\
    \\
    You are a helpful assistant. \\
    \tcblower
    
    \textbf{{\userRelationship}:}\\
    \\
    \textcolor{blue}{\texttt{\{demographic sentence\}}} \textcolor{blue}{\texttt{\{persona sentence\}}} So, I <relationship>.\\
    \\
    Generate the most appropriate sentence for "<relationship>" in the given sentence. You must provide the answer corresponding to "<relationship>".\\
    <relationship>: 
\end{prompt}

\begin{prompt}{Prompt Template for Social Persona Commonsense Generation: \Experience}
    \textbf{{\system}:}\\
    \\
    You are a helpful assistant. \\
    \tcblower
    
    \textbf{{\userExperience}:}\\
    \\
    I <experience>. Now, \textcolor{blue}{\texttt{\{demographic sentence\}}} \textcolor{blue}{\texttt{\{persona sentence\}}} \\
    \\
    Generate the most appropriate sentence for "<experience>" in the given sentence. You must provide the answer corresponding to "<experience>". \\
    <experience>: 
\end{prompt}

\begin{prompt}{Prompt Template for Social Persona Commonsense Generation: \Characteristic}
    \textbf{{\system}:}\\
    \\
    You are a helpful assistant. \\
    \tcblower
    
    \textbf{{\userCharacteristic}:}\\
    \\
    \textcolor{blue}{\texttt{\{demographic sentence\}}} \textcolor{blue}{\texttt{\{persona sentence\}}} I <characteristic>. \\
    \\    
    Generate the most appropriate sentence for "<characteristic>" in the given sentence. You must provide the answer corresponding to "<characteristic>".\\
    <characteristic>: 
\end{prompt}

\begin{prompt}{Prompt Template for Social Narrative Generation}
    \textbf{{\system}:}\\
    \\
    You are a helpful assistant.
    \\    
    \tcblower
    
    \textbf{{\user}:}\\
    \\
    \textcolor{blue}{\texttt{\{narrative sentence form\}}} \\
    \\
    Rewrite this sentence with more specific details in two or three sentences: 
\end{prompt}

\begin{prompt}{Prompt Template for Social Event Graph Generation}
    \textbf{{\system}:}\\
    \\
    You should generate a temporal event graph composed of daily events occuring in a person's life. The temporal event graph contains nodes and edges. Each node represents a daily event which is written in two or three sentences. Each edge represents the casual relationship between two nodes (events), i.e., a past event -> current event. The current event is determined by how much time has passed since the past event and what personal experiences were had during that period. You must generate the temporal event graph following the guidelines below.\\
    \\
    \lbrack Guideline\rbrack \\
    - The graph is represented in the form of a json list.\\
    - Each entry is a python dictionary containing the following keys: "id", "event", "date", "caused\_by".\\
    - The "id" field contains a unique identifier for the current event.\\
    - The "event" field contains a description of the current event.\\
    - The "date" field contains a specific date of the current event and is represented in the form of "\%Y.\%m.\%d".\\
    - The "caused\_by" field represents the edge (i.e., a past event) and is represented in the form of a python dictionary containing the following keys: "caused\_by:id", "caused\_by:time\_interval", "caused\_by:experience\_op", "caused\_by:experience".\\
    - The "caused\_by:id" field contains an "id" of the past event that has caused the current event.\\
    - The "caused\_by:time\_interval" field contains a time interval between the past event and the current event.\\
    - The "caused\_by:experience\_op" field contains an episodic experience operation.\\
    - The "caused\_by:experience" field contains a short description of the added or updated episodic experience.\\
    - The unit of time interval is \lbrack "hour", "day", "week", "month", "year"\rbrack.\\
    - The selected time interval should be formatted as "<base number> <time interval unit>".\\
    - List of the episodic experience operation is \lbrack "add", "update"\rbrack.\\
    - The "add" operation refers to an operation that adds a new experience that have not been encountered in the past.\\
    - The "update" operation refers to an operation that updates a past experience with a new experience.\\
    - Events/Experiences can be positive or negative events or experiences.\\
    - Events in the "caused\_by:id" field should occur on dates before the current event that they have caused.\\
    - If there is no entry of "caused\_by" field, then you should generate an empty dictionary.
    - Each event must be written in the present tense.\\
    - The year in the "date" field must be until April 2024. \\
    - You should generate the temporal event graph based on commonsense or a world model.\\
    
    \tcblower
    
    \textbf{{\user}:}\\
    \\
    \textcolor{blue}{\texttt{\{name\}}}'s initial personal event: \textcolor{blue}{\texttt{\{event\}}}\\
    \\
    Given the \textcolor{blue}{\texttt{\{name\}}}'s initial personal event, generate the temporal event graph containing more than five events.\\
    Temporal Event Graph: 
\end{prompt}

\begin{prompt}{Prompt Template for Device-Stored Image Description Generation}
    \textbf{{\system}:}\\
    \\
    Given the sentence related to a person's daily life, your task is to generate five image descriptions that could be stored on the person's mobile device, along with corresponding image categories. You should use the format "<image\_description> (Category: <image\_category>)". The image category may include selfies, past memories, screenshots, landmarks, animals, art, celebrities, nature, and food.\\
    \\
    For example, \\
    \\
    My name is Tom. I am a 32-year-old man. I was born in the USA and currently reside there. I have a strong interest in basketball. I played basketball in middle school, but now I work as a chatbot developer at a startup. I enjoy watching the NBA because I love basketball.\\
    \\
    Image descriptions stored on Tom's mobile device:\\
    1. A photo of a young Tom playing basketball in a middle school gymnasium (Category: Past Memory, Sport)\\
    2. A selfie of Tom smiling at the Golden State Warriors' arena during a game (Category: Selfie, Sport)\\
    3. A screenshot of chatbot development code using Python (Category: Screenshot, Computer, Software)\\
    4. A picture of Tom enjoying a night out with coworkers at a local pub (Category: Social Networking, Food, Drink)\\
    5. A photo of Tom meeting a famous NBA player at a basketball event (Category: Celebrity, Sport)\\
    \tcblower
    
    \textbf{{\user}:}\\
    \\
    \textcolor{blue}{\texttt{\{narrative\}}}\\
    \\
    Given the sentence above, generate five possible image descriptions that are stored on \textcolor{blue}{\texttt{\{name\}}}'s mobile device. For example, images may include selfies, past memories, screenshots, landmarks, animals, art, celebrities, nature, and food.\\
    1. 
    
\end{prompt}

\begin{prompt}{Prompt Template for Social Multi-Modal Dialogue Generation}
    \textbf{{\system}:}\\
    \\
    Your job is to generate a long in-depth conversation between an user and an user-friendly AI assistant with multiple turns. The user and AI assistant can share images during a conversation in order to strengthen social relationship, to convey important information, to amuse/entertain, to clarify complex situations, to change the topic of dialogue, or to express emotions/opinions/reactions. There must be more than two image-sharing moments within the conversation. The shared images can either be from the collection previously stored on the user's mobile device or obtained from the internet. You must generate the conversation following the guidelines below.\\
    \\
    \lbrack Guideline\rbrack \\
    - The conversation is represented in the form of a json list.\\
    - Each entry is a python dictionary containing the following keys: "utterance\_id", "speaker", "utterance", "sharing\_info".\\
    - The "utterance\_id" field contains a unique identifier for the utterance within the conversation.\\
    - The "speaker" field contains a speaker of the utterance.\\
    - The "utterance" field contains the utterance of the speaker. If the image-sharing behavior occurs, then the "utterance" is a empty string.\\
    - The "sharing\_info" field represents the image-sharing moment and is represented in the form of a python dictionary containing the following keys: "rationale", "image\_description", "image\_source", "keywords", "image\_id\_from\_mobile".\\
    - If the image-sharing moment does not occur, then the "sharing\_info" field is an empty python dictionary.\\
    - The "rationale" field represents the reason behind sharing the relevant image.\\
    - The "image\_description" field contains a description of the shared image.\\
    - The "image\_source" field contains a source of the shared image whether it is from the internet (internet) or the user's mobile device (mobile).\\
    - If you select the user's mobile device as the "image\_source," you must either share an image that matches one of the existing descriptions already on the user's mobile device or share a new image that does not exist among these descriptions.\\
    - If you share an image that matches one of the existing descriptions on the user's mobile device, you must generate the appropriate image ID in the "image\_id\_from\_mobile" field.\\
    - If you share a new image that does not match any existing descriptions on the user's mobile device, you must enter "new added image" in the "image\_id\_from\_mobile" field.\\
    - The "keywords" field contains keywords of the shared image.\\
\end{prompt}

\begin{prompt}{Prompt Template for First Round Social Multi-Modal Dialogue Generation}
    \textbf{{\user}:}\\
    \\
    \textcolor{blue}{\texttt{\{name\}}}'s Profile Information: \\
    - Age: \textcolor{blue}{\texttt{\{age\}}} \\
    - Gender: \textcolor{blue}{\texttt{\{gender\}}} \\
    - Birthplace: \textcolor{blue}{\texttt{\{birthplace\}}} \\
    - Residence: \textcolor{blue}{\texttt{\{residence\}}} \\
    \\
    Existing image descriptions in \textcolor{blue}{\texttt{\{name\}}}'s mobile device: \textcolor{blue}{\texttt{\{device-stored image descriptions\}}} \\
    \\
    The topic of the conversation between the AI assistant and \textcolor{blue}{\texttt{\{name\}}} on \textcolor{blue}{\texttt{\{date\}}} today is as follows. \\
    - Topic on \textcolor{blue}{\texttt{\{date\}}}: \textcolor{blue}{\texttt{\{event\}}} \\
    \\
    Generate a long, in-depth conversation with multiple turns based on the given {name}'s profile information and the current topic of conversation. \\
    \\
\end{prompt}

\begin{prompt}{Prompt Template for N-th Round Social Multi-Modal Dialogue Generation}
    \textbf{{\user}:}\\
    \\
    \textcolor{blue}{\texttt{\{name\}}}'s Profile Information: \\
    - Age: \textcolor{blue}{\texttt{\{age\}}} \\
    - Gender: \textcolor{blue}{\texttt{\{gender\}}} \\
    - Birthplace: \textcolor{blue}{\texttt{\{birthplace\}}} \\
    - Residence: \textcolor{blue}{\texttt{\{residence\}}} \\
    \\
    Existing image descriptions in \textcolor{blue}{\texttt{\{name\}}}'s mobile device: \textcolor{blue}{\texttt{\{device-stored image descriptions\}}} \\
    \\
    The topics of the conversation the user had with AI assistant by date are as follows: \\
    \textcolor{blue}{\texttt{\{event history\}}} \\
    \\
    \textcolor{blue}{\texttt{\{time interval\}}} later from the \textcolor{blue}{\texttt{\{last date\}}}, on \textcolor{blue}{\texttt{\{date\}}} today, \textcolor{blue}{\texttt{\{name\}}} has gone through a new experience, and based on this experience, \textcolor{blue}{\texttt{\{name\}}} and the AI assistant engage in a conversation today. The new experience \textcolor{blue}{\texttt{\{name\}}} went through and the topic of conversation with the AI assistant are as follows. \\
    - \textcolor{blue}{\texttt{\{name\}}}'s Experience: \textcolor{blue}{\texttt{\{experience\}}} \\
    - Topic on \textcolor{blue}{\texttt{\{date\}}}: \textcolor{blue}{\texttt{\{event\}}} \\
    \\
    Generate a long, in-depth conversation with multiple turns based on the given \textcolor{blue}{\texttt{\{name\}}}'s profile information, the last topic of conversation, the experience and the current topic of conversation.\\
    \\
\end{prompt}

\begin{prompt}{Prompt Template for \planExecute Generation}
    \textbf{{\system}:}\\
    \\
    Your job is to determine the most appropriate module from a list of models to process the input request. Please select one module from the following list: \\
    \\
    Personalized Text-to-Image Generator: This module generates personalized images from a given description and a human face image. For example, if you provide a face image and a description like “A selfie of Tom smiling at the Golden State Warriors' arena during a game,” the module will generate a customized realistic human image. Note that when you generate the answer, you must generate the module name and modified image description together. The modified image description MUST include a strict format: “<class\_word> \lbrack img\rbrack”. <class\_word> represents the identity of a human, such as a man, woman, girl, boy, or young boy, etc. \lbrack img\rbrack denotes the special token. You must not omit this strict format, and you must keep the original image description as it is and only add this strict format. \\
    \\
    Web Search: This module finds related images from the internet in real-time based on the given user's input image description. The image description is primarily related to the latest information. Therefore, this method is useful when up-to-date information is needed. \\
    \\
    Image Database Retrieval: This module finds relevant images from a pre-built image database based on the given user's input image description. To build an image database containing images on various topics, images are collected from the RedCaps, Conceptual Captions 12M (CC12M), ChartQA, AI2D, and MathVision datasets. Descriptions related to each dataset are as follows: \\
    - RedCaps: This is a large-scale dataset of 12M image-text pairs collected from Reddit. Images and captions from Reddit depict and describe a wide variety of objects and scenes. \\
    - CC12M: This is a dataset with 12 million image-text pairs specifically meant to be used for vision and language pre-training. It is larger and covers a much more diverse set of visual concepts than the Conceptual Captions (CC3M). \\
    - ChartQA: This is a large-scale ChartQA dataset with real-world charts and human-authored question-answer pairs. This dataset covers 9.6K chart images. \\
    - AI2D: This is a dataset of over 5,000 grade school science diagrams with over 150,000 rich annotations, their ground truth syntactic parses, and more than 15,000 corresponding multiple choice questions. \\
    - MathVision: This dataset is a meticulously curated collection of 3,040 high-quality mathematical problems with visual contexts sourced from real math competitions. Spanning 16 distinct mathematical disciplines and graded across 5 levels of difficulty. \\
    \\
    For example, \\
    \\
    Name: Tom \\
    Gender: Male \\
    Age: 21 \\
    Image Description: A selfie of Tom smiling at the Golden State Warriors' arena during a game \\
    Module: Personalized Text-to-Image Generator \\
    Modified Image Description: A selfie of a young man [img] smiling at the Golden State Warriors' arena during a game \\
    \\
    Name: Tom \\
    Gender: Male \\
    Age: 21 \\
    Image Description: A screenshot of chatbot development code using Python \\
    Module: Image Database Retrieval \\
    \\
    Name: Tom \\
    Gender: Male \\
    Age: 21 \\
    Image Description: A photo of Manchester United lifting the 2023-24 FA Cup trophy \\
    Module: Web Search \\
    \tcblower
    
    \textbf{{\user}:}\\
    \\
    Name: \textcolor{blue}{\texttt{\{name\}}} \\
    Gender: \textcolor{blue}{\texttt{\{gender\}}} \\
    Age: \textcolor{blue}{\texttt{\{age\}}} \\
    Image Description: \textcolor{blue}{\texttt{\{image description\}}} \\
    Module: \\
\end{prompt}

\begin{prompt}{Prompt Template for First Round Dialogue Summarization Generation}
    \textbf{{\system}:}\\
    \\
    Your job is to summarize the given conversation. \\
    \tcblower
    
    \textbf{{\user}:}\\
    \\
    The conversation between \textcolor{blue}{\texttt{\{name\}}} and the AI assistant on \textcolor{blue}{\texttt{\{current\_date\}}} today is as follow.\\
    \\
    \textcolor{blue}{\texttt{\{dialogue\}}}\\
    \\
    Summarize the given conversation between \textcolor{blue}{\texttt{\{name\}}} and the AI assistant so far. Include key details about both speakers and include time references.\\
    Summarization: 
    \\
\end{prompt}

\begin{prompt}{Prompt Template for N-th Round Dialogue Summarization Generation}
    \textbf{{\system}:}\\
    \\
    Your job is to summarize the given conversation. \\
    \tcblower
    
    \textbf{{\user}:}\\
    \\
    In the previous interaction, \textcolor{blue}{\texttt{\{previous\_summary\}}}. \textcolor{blue}{\texttt{\{time\_interval\}}} later from the \textcolor{blue}{\texttt{\{last\_date\}}}, the conversation between \textcolor{blue}{\texttt{\{name\}}} and the AI assistant on {\texttt{\{current\_date\}}} today is as follow. \\
    \\
    \textcolor{blue}{\texttt{\{dialogue\}}}\\
    \\
    Summarize the given conversation between \textcolor{blue}{\texttt{\{name\}}} and the AI assistant so far. Include key details about both speakers and include time references. \\
    Summarization: 
    \\
\end{prompt}

\end{document}